%% file: bare_jrnl_new_sample4.tex
\documentclass[lettersize,journal]{IEEEtran}
\usepackage{amsmath,amsfonts}
\usepackage{algorithm}
\usepackage{array}
\usepackage[caption=false,font=normalsize,labelfont=sf,textfont=sf]{subfig}
\usepackage{textcomp}
\usepackage{stfloats}
\usepackage{url}
\usepackage{verbatim}
\usepackage{graphicx}
\usepackage{cite}
\usepackage{CJKutf8}  

\usepackage[noend]{algpseudocode}
\usepackage{tikz}
\usetikzlibrary{shapes.geometric} 
\usepackage{bm} 
\usepackage{amssymb}
\usepackage{pgfplots}
\pgfplotsset{width=10cm,compat=1.9}

\usepackage{hyperref}

\usepackage{enumitem} 

\usepackage[T1]{fontenc}

\usepackage[utf8]{inputenc}

\usepackage{microtype}

\usepackage{inconsolata}

\usepackage[most]{tcolorbox}
\usepackage{alltt}
\global

\definecolor{aigold}{RGB}{244,210, 1} 
\definecolor{aigreen}{RGB}{210,244,211}

\definecolor{aired}{RGB}{255,180,181}

\tcbset{
  aibox/.style={
    width=474.18663pt,
    top=10pt,
    colback=white,
    colframe=black,
    colbacktitle=black,
    enhanced,
    center,
    attach boxed title to top left={yshift=-0.1in,xshift=0.15in},
    boxed title style={boxrule=0pt,colframe=white,},
  }
}
\newtcolorbox{AIbox}[2][]{aibox,title=#2,#1}

\begin{document}

\title{ClarQ-LLM: A Benchmark for Models Clarifying and Requesting Information in Task-Oriented Dialog}

\author{Yujian Gan, Changling Li, Jinxia Xie, Luou Wen, Matthew Purver, Massimo Poesio
\thanks{Yujian Gan, Matthew Purver and Massimo Poesio are with Queen Mary University of London, Mile End Road, London E1 4NS, UK (e-mail: y.gan@qmul.ac.uk; m.purver@qmul.ac.uk; m.poesio@qmul.ac.uk); Purver is also affiliated to the Jožef Stefan Institute, Ljubljana, Slovenia; and Poesio is also affiliated to the University of Utrecht, Utrecht, Netherlands.}
\thanks{Changling Li and Jinxia Xie are with the Guangxi Normal University, 15 Yucai Road,  Guilin 541004, China (e-mail: lichangling@stu.gxnu.edu.cn, jinxia\_xie@hotmail.com).}

}

\markboth{Journal of \LaTeX\ Class Files,~Vol.~14, No.~8, August~2021}%
{Shell \MakeLowercase{\textit{et al.}}: A Sample Article Using IEEEtran.cls for IEEE Journals}

\IEEEpubid{This work has been submitted to the IEEE for possible publication. Copyright may be transferred without notice, after which this version may no longer be accessible.}

\maketitle

\begin{abstract}
We introduce ClarQ-LLM, an evaluation framework consisting of bilingual English-Chinese conversation tasks, conversational agents and evaluation metrics, designed to serve as a strong benchmark for assessing agents' ability to ask clarification questions in task-oriented dialogues.
The benchmark includes 31 different task types, each with 10 unique dialogue scenarios between information seeker and provider agents. 
The scenarios require the seeker  to ask questions to resolve uncertainty and gather necessary information to complete tasks.
Unlike traditional benchmarks that evaluate agents based on fixed dialogue content, ClarQ-LLM includes a provider conversational agent to replicate the original human provider in the benchmark. 
This allows both current and future seeker agents to test their ability to complete information gathering tasks through dialogue by directly interacting with our provider agent.
In tests, LLAMA3.1 405B seeker agent managed a maximum success rate of only 60.05\%, showing that ClarQ-LLM presents a strong challenge for future research
\footnote{You can find the benchmarks \href{https://github.com/ygan/ClarQ-LLM}{here}.}
\end{abstract}


\input{intro}

\input{related_work}

\input{dataset}

\input{helper}

\input{evaluation}
\input{experiment}
\input{Conclusion}

\bibliography{custom}
\bibliographystyle{IEEEtran}

\input{appendix/two-mode}
\input{appendix/data-examples}
\input{appendix/conversation_example}


 




\vfill

\end{document}

%% file: intro.tex
\section{Introduction}

\begin{figure}[t]
    \includegraphics[width=0.48\textwidth]{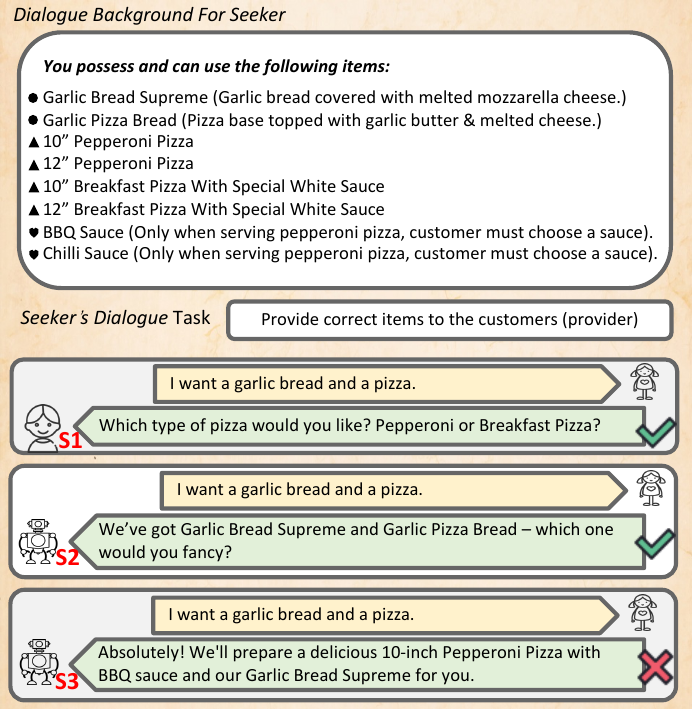}
    \centering
    \caption{An example of a dialogue background and task.  In this task, the provider engages in dialogue with three seekers: S1, S2, and S3. Responses from S1 and S2, which both clarify required but previously unspecified information in different ways, are considered acceptable. However, response S3, which assumes uncertain knowledge, is not acceptable.
    }
    \label{figure:conv}
  \end{figure}

Task-oriented dialogues have always been a hot topic in the research on dialogue systems~\cite{880078,balaraman-etal-2021-recent,Qin2023}. 
The typical scenario for these dialogues involves an interaction between a 
helper agent that needs to acquire information in order to help
(referred to as `{\bf (information) seeker}' for simplicity) 
and a user who interacts with the helper/seeker to carry out a task 
(referred to as `{\bf (information) provider}'). 
In this setting, the seeker must engage in conversation with the provider to gather the required information to complete a task, such as making hotel reservations~\cite{Budzianowski2018}, navigation~\cite{eric2017keyvalue}, and others.
This process often requires asking clarification questions~\footnote{The definition of the clarification question here follows\cite{Li-Changling-2024-Analyzing}, encompassing both information-seeking and clarification related to uncertainty.}, due to the potentially complex nature of these tasks, the seeker's possible lack of crucial details about them and/or the ambiguity or incompleteness of the input offered by the provider
(see e.g.~\cite{madureira-schlangen-2023-instruction}).

However, the availability of benchmarks specifically designed for task-oriented information seeking with clarification questions is limited. 
Among the few examples are ShARC and its variants~\cite{Saeidi2018,Feng2023}.
These benchmarks primarily contain clarifying questions with responses limited to ``yes'' or ``no'', and the knowledge points needing clarification are listed in advance, making the dialogue task appear relatively simple by current standards. 
They also 
rely mostly on evaluation metrics like BLEU~\cite{papineni-etal-2002-bleu}, requiring a model's output to closely resemble the annotated data in its surface form rather than checking its semantic content or intent, and requiring it to clarify information in a particular sequence rather than allowing more flexibility.

\IEEEpubidadjcol

\begin{figure}[t]
    \includegraphics[width=0.48\textwidth]{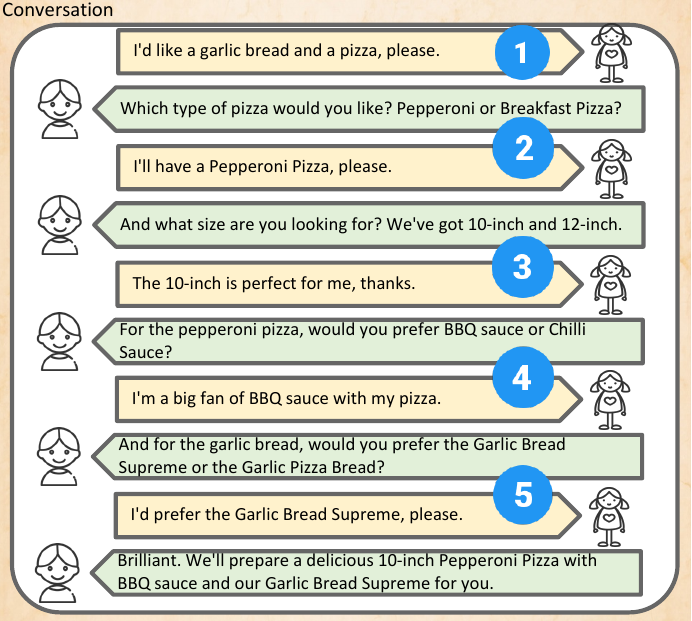}
    \centering
    \caption{The complete dialogue of the task in Figure~\ref{figure:conv}.}
    \label{figure:conv-2}
  \end{figure}

To 
address 
more complex clarification behaviour, and evaluate 
such systems 
in a more functional way,  we present ClarQ-LLM, an evaluation framework consisting of different dialogue scenarios, evaluation metrics and an information provider agent powered by GPT-4o or LLAMA3.1-405B. Each scenario is annotated with the information required for task completion; seeker agents must interact with the provider to obtain this information via clarification, and are evaluated on their success in doing this.
Figure~\ref{figure:conv} provides a scenario example from ClarQ-LLM,
while Figure~\ref{figure:conv-2} presents a conversation based on this scenario between the seeker (on the left in the image) and the provider (on the right). 

ClarQ-LLM requires 
seekers to ask clarification questions to accurately understand the provider's intentions:
it is designed to assess the ability of LLM seeker agents to identify uncertainties in dialogues and eliminate them through clarification questions.
As LLMs have evolved to generate fluent and precise language, persisting in the use of traditional fluency and match-based evaluation metrics to assess their context-specific responses appears increasingly inadequate.
For example, as illustrated in Figure~\ref{figure:conv}, the response from S2 
to the same sentence significantly differs from that of the human seeker (S1).
When evaluated using traditional metrics such as BLEU or dialogue state tracking, the S2 seeker agent's response would likely be assessed negatively.
Nevertheless, ClarQ-LLM contends that the S2 seeker agent's response is not problematic, despite its complete divergence from the human seeker's. 
Since the customer did not specify the types of pizza and bread desired, inquiring about the type of pizza or bread first seems acceptable.

ClarQ-LLM utilizes the success rate of the dialogue tasks as one of the metrics to evaluate the seeker. 
The success of a dialogue task hinges on the seeker agent’s ability to gather all necessary information from the provider agent.
For instance, in the dialogue tasks depicted in Figures~\ref{figure:conv} and~\ref{figure:conv-2}, the seeker must obtain crucial details from the provider agent, such as the pizza's type and size, as well as the bread type. 
The human seeker in Figures~\ref{figure:conv-2} successfully obtains these 
by asking several questions.
Conversely, even though the response of the S3 seeker agent in Figure~\ref{figure:conv} correctly guesses the customer's needs, it is still considered incorrect. 
Because the customer in Figure~\ref{figure:conv} did not specify the exact items desired, the seeker agent should not make assumptions but should instead ask clarification questions to understand the customer's true needs. 
Even if the seeker agent in this example fortuitously guessed correctly, reliance on such guesswork cannot ensure consistently accurate responses; thus, ClarQ-LLM deems it incorrect. 

The provider agent in ClarQ-LLM can capture the genuine needs of human providers and simulate conversations with the seeker agent, acting as human providers would.
This ensures that regardless of whether the seeker agent inquires about bread or pizza first, the provider agent can deliver appropriate responses. 
Upon concluding the dialogue, we determine the success of the seeker agent by checking if it has gathered all necessary information from the provider agent.
Experimental results indicate that, even with the use of state-of-the-art models like GPT-4o and LLAMA3.1-405B, the success rates of LLM-based seekers on ClarQ-LLM are only 50.8\% and 60.5\%, respectively. These figures are significantly lower than the 85\% success rate achieved by human seekers.
This indicates that our tasks pose a significant challenge for future research.

In summary, this paper discusses two types of agents: the seeker agent and the provider agent. ClarQ-LLM includes  different conversational tasks, evaluation metrics, and provider agents based on GPT-4o or LLAMA3.1-405B. ClarQ-LLM is designed primarily to evaluate the seeker agent, not the provider agent: the goal is to assess whether seeker agents can identify uncertainties in their dialogues with provider agents and 
pose relevant clarification questions accordingly.

%% file: related_work.tex
\section{Related Work}
\label{sec:Related Work}
\subsection{Task-Oriented Dialogues}

Task-oriented dialogue focuses on conducting conversations to complete a specific task. 
Some task-oriented dialogues are accompanied by a task of dialogue state tracking, which requires the model to understand the user's needs and update the dialogue state as the conversation progresses, while also generating reasonable natural language responses to the user~\cite{williams-etal-2013-dialog,Dai2021}.
With the advancement of dialogue system technology, researchers have shifted their focus from previous single-domain dialogue datasets to multi-domain task-oriented dialogue datasets~\cite{Wen2017,Budzianowski2018,eric-etal-2020-multiwoz,quan-etal-2020-risawoz}.

However, 
there are not many 
task-oriented dialogue datasets focusing on clarification questions%
~\cite{Rahmani2023}.
The recent datasets ShARC~\cite{Saeidi2018} and ClariT~\cite{Feng2023} contain clarification questions that only require a "yes" or "no" response, and the knowledge points needing clarification are usually listed in advance, making the dialogue task appear relatively simple. 
Beyond the domain of task-oriented dialogue, Aliannejadi et al.\cite{aliannejadi-etal-2021-building} propose a dataset and task for clarification in chat. However, it is not suitable for our purposes, as it focuses solely on information-seeking.
Therefore, we believe it is crucial to introduce a more challenging dataset that, like earlier datasets, facilitates the gathering of needed information to accomplish tasks through the process of asking clarification questions.
What sets it apart is the flexibility in responses to these clarification questions, alongside the model's need to be able to ask other types of questions for more information.

\subsection{Clarification Question in Dialogues}

Clarification questions in conversations help figure out what someone really means. 
A common scenario is serving users' information query needs, where the system uses clarification questions to grasp the user's intent and then provides the appropriate information~\cite{rao-daume-iii-2018-learning,rao-daume-iii-2019-answer,xu-etal-2019-asking,10.1145/3331184.3331265,zamani2020mimics}. 
However, we have noticed that LLMs like GPT-4, ChatGPT, and BARD often skip asking these clarification questions when helping users find information.

Take this scenario mentioned in~\cite{aliannejadi-etal-2021-building}: a user asks, "Tell me about the source of the Nile." 
The system should pose a clarification question to the user, asking whether they are interested in learning about the source of the river Nile or the board game ``Source of the Nile'' \cite{aliannejadi-etal-2021-building}.
Tests with ChatGPT show that it tends to directly introduce the origins of the Nile River, even if the user may be interested in the board game.
However, this is not necessarily a significant drawback: ChatGPT can still catch up and provide the correct information after the user clarifies that they meant the board game, with just an additional sentence.
Skipping clarification questions doesn't necessarily complicate the process for users: some end up getting what they want after speaking the same sentences anyway; while others actually looking for information on the Nile River get their answer faster, since they don't have to answer clarification questions.

This shows that, even though ChatGPT did not ask clarification questions, it can still give a good user experience. 
Inspired by this, ClarQ-LLM allow the model to gather information through both clarification and non-clarification questions, and we opted not to use common metrics such as BLEU~\cite{papineni-etal-2002-bleu}, ROUGE~\cite{lin-2004-rouge}, and METEOR~\cite{banerjee-lavie-2005-meteor} that evaluate on the basis of an assumed ideal gold standard.

%% file: dataset.tex
\section{Conversation Task Construction}

\begin{figure}[t]
    \includegraphics[width=0.48\textwidth]{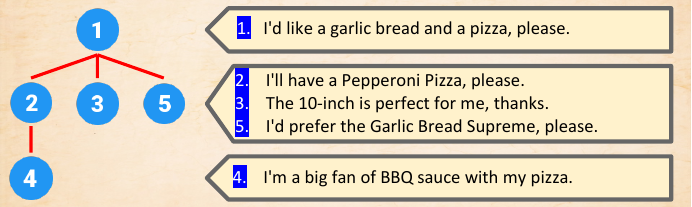}
    \centering
    \caption{The provider response tree extracted from Figure~\ref{figure:conv-2}.}
    \label{figure:tree}
  \end{figure}

\subsection{Overview}

ClarQ-LLM encompasses a diverse range of dialogue tasks, each accompanied by its background and conversations between a human seeker and provider.
All these tasks were written by four PhDs and graduate students over a total of 700 person-hours. 
In the ClarQ-LLM conversations, the provider intentionally uses language filled with uncertainties based on the task background, prompting each conversational task to contain multiple pieces of information that seeker agents must clarify through clarification questions.
To ensure diversity, we have designed 31 different types of tasks, including common travel planning, specialized electrical engineering, and even fictional magical dialogues.
For each type of task, there are 10 unique task instantiations, culminating in a total of 310 task instantiations\footnote{All our task types can be found \href{https://github.com/ygan/ClarQ-LLM}{here}.}.
We've organized these 31 types into two main groups: 26 for testing, to evaluate how well seekers perform, and 5 for development, aimed at building and testing applications. 
ClarQ-LLM doesn't include a training set.

Regarding the background, ClarQ-LLM includes a diverse range of 392 items, 341 different skills or actions, and 736 unique scenes. 
For the quantity of uncertainty, our tasks encompass a variety of setups: 1 task comes with 2 uncertainties, 56 tasks with 3 uncertainties, 123 tasks with 4 uncertainties, 95 tasks with 5 uncertainties, and 35 tasks with 6 uncertainties. 
In terms of structural complexity, as measured by the depth of provider response trees, ClarQ-LLM includes 166 tasks structured within trees of depth 1, 124 tasks structured within trees of depth 2, 18 tasks structured within trees of depth 3, and only 2 tasks that feature the most complex structures within trees of depth 4.

\subsection{Construction Principles}
The conversation tasks we constructed follows three basic principles to ensure the quality and usability.

\paragraph{Question types}

While there are various causes and types of clarification question
(see e.g. \cite{rodriguez-schlangen-2004-form,rieser-moore-2005-implications,purver-etal-2001-means}),
our ClarQ-LLM's primary objective is to determine if the model can effectively ask relevant questions about ambiguous references. 
However, merely asking these clarification questions does not suffice to gather all the necessary information for completing a task. 

\paragraph{Talking in a Virtual Game World}
We carry out our dialogue tasks in different virtual game worlds that we've designed with a lot of care. 
These worlds might have magical elements, like casting spells or fighting monsters.
Even though these tasks happen in a game setting, some of them are closely linked to real-life skills. 
For instance, tasks related to electronic engineering require the player to have knowledge of electronic parts. This approach tests whether the model can adapt to entirely new tasks and possess specific real-world knowledge.

\paragraph{Following Ethical Rules}

When creating our database, we're very careful about ethics, avoiding any content that could lead to ethical issues. For example, we use fictional monsters instead of involving tasks that could harm people or animals. Likewise, for cooking tasks, we opt for vegetarian options to prevent harming animals.



\subsection{Human Provider Response Tree}
\label{sec:Provider Response Tree}
ClarQ-LLM encompasses not only dialogue tasks, backgrounds, and content but also annotates human provider responses in a tree structure. 
This helps provider agents 
to 
more accurately simulate human providers from the original task. 
Figure~\ref{figure:tree} displays the tree-structured annotations of provider responses for the dialogue tasks in Figures~\ref{figure:conv} and~\ref{figure:conv-2}. 
This top-down tree structure clearly delineates the logical sequence in which the provider should offer information to the seeker.
For example, the tree indicates that the provider initially informs the seeker of the information at node 1, which introduces three uncertainties: the type and size of the pizza, and the type of bread, with uncertainties resolved by provider responses at nodes 2, 3, and 5, respectively. 
This means that the seeker needs to analyze the uncertainties in the content of node 1 and pose clarification questions accordingly. 
Since the uncertainty at node 4 is related to node 2, we place node 4 after node 2. 
This ensures that the dialogue for node 4, which involves determining the sauce type, only occurs after node 2 where Pepperoni Pizza is mentioned. 
This stipulation is because a breakfast pizza does not require choosing a sauce, as detailed in the conversation background in Figure~\ref{figure:conv}. 

It's important to note that this tree is intended primarily to ensure the logical correctness of the provider; it does not impose any constraints on the seeker. 
The seeker is free to question the provider based on their own judgment.
For example, some seekers might overlook the information that the provider requested a pizza, and then ask, ``How many pizzas do you need?'' 
In this situation, the provider agent will inform the seeker again.
For details about how the provider agent works with this tree, please refer to Section~\ref{sec:Provider Agent}.

\subsection{Annotation Steps}

We develop all tasks in five steps, involving around 700 hours of human labor in total:

Step 1: Creating Initial Dialogues Between the Seeker and Provider. We assembled a team consisting of four college students tasked with writing the dialogues in Chinese. The diverse composition of the team contributed to a greater variety of tasks.

Step 2: Improving the Drafts. After completing the initial versions, we identified many errors and illogical parts. A post-doctoral researcher meticulously reviewed each task to correct any issues, ensuring that everything was logical and adhered to the principles of data annotation. Simultaneously, he annotated the provider's responses into a tree.

Step 3: Reviewing the Chinese Version. Next, another undergraduate checked the Chinese data for anything that might be missing or unclear, didn't make sense, contained typos, or whether the provider response tree was correctly organized.

Step 4: Translation and Cultural Adaptation. After finalizing the Chinese version, we utilized GPT-4 to translate it into English. A bilingual student reviewed the translation to correct any inaccuracies. Then, a PhD student, who is a native English speaker, evaluated the dialogues to ensure they were clear, logical, and consistent with Western cultural expectations.

Step 5: Testing with Real Conversations. Once everything was reviewed and fixed, we asked some students who hadn’t seen the data before to talk with the provider agent model. This was to see how well the model worked and if the way we organized the provider's responses were right. After these tests, we had our final conversation tasks.

%% file: helper.tex
\section{Provider Agent}
\label{sec:Provider Agent}
\subsection{Overview}
To provide future seeker models with a dialogue partner without relying on human providers, we developed a provider agent based on the state-of-the-art models GPT-4o or LLAMA3.1-405B.
This provider agent can simulate human providers, understand the seekers' language, and provide appropriate responses.
Unlike seeker agents, who can talk freely, the provider agent aims to replicate the human provider in ClarQ-LLM as closely as possible.
To this end, we have marked the responses of the human provider as tree structures to help the provider agent replicate the needs, language, and logic of the human provider.

For example, based on the tree in Figure~\ref{figure:tree}, the provider agent will first inform the seeker of the content of node 1 just like the human provider would: ``I'd like a garlic bread and a pizza, please.'' 
In this way, the provider agent introduces the same three uncertainties as the human provider: the type and size of the pizza, and the type of bread.
Afterwards, the seeker can generate its questions based on its analysis.
If the question posed by the seeker can be answered using the content of nodes 2, 3, and 5, the provider agent will generate responses based on the content of these nodes.
However, the seeker may illogically asks the provider agent after the provider's node 1 response: `Would you prefer BBQ sauce or Chilli Sauce?'\footnote{Firstly, the provider (customer) did not mention wanting sauce, and secondly, this question should be asked after confirming that the customer wants a pepperoni pizza, because a breakfast pizza does not require choosing a sauce.} or even pose an unrelated question like `How's the weather today?'. 
Faced with such questions, the provider agent will automatically generate a sensible reply to alleviate the seeker agent's concerns and steer them back to the dialogue task (this will be further discussed in the next subsection).

\subsection{Ensuring Valid Responses}
\label{sec:Ensuring Valid Responses}

The provider agent cannot freely generate responses; otherwise, it may produce invalid responses, leading to unfair comparisons between seeker models.
To prevent the generation of invalid sentences, we have imposed restrictions on the response generation.
First, we categorize the responses into two types: \textbf{Response I} (responses from the human provider response tree) and \textbf{Response II} (responses generated based on the conversation context).

For \textbf{Response I}, which contains information necessary for completing the task, we require that \textbf{Response I} retain the original text from the human provider.
However, additional content can be added before or after the original text to make the conversation flow more smoothly. 
Since the sentences generated by the agent include the human's original text, this ensures that the provider delivers accurate and complete information to the seeker, and all seekers receive consistent original information.
For \textbf{Response II}, which does not contain information necessary for completing the task, this type of response does not affect the judgment of task success. 
The provider agent only needs to generate \textbf{Response II} when \textbf{Response I} cannot address the seeker’s request.
For example, in the conversation shown in Figure~\ref{figure:conv}, suppose the seeker asks: ``A 10-inch pizza is quite large; I'm worried you won't finish it. Are you sure you want to order it?'' 
For this task-irrelevant question, we cannot find an appropriate response in the human provider text, so the provider will generate a reasonable response based on the previous conversation with the seeker, reassuring the seeker and guiding them back to the task.
The \textbf{Response I} ensures the consistency of information, while the \textbf{Response II} ensures that the seeker is guided back to the conversation task if they stray. 
Together, they ensure that different seeker agents receive fair evaluations.

\subsection{Implementation Details}
\label{sec:Implementation of the Provider Agent}

\begin{algorithm*}[t]
  \caption{Generate provider agent response }\label{alg:provider}
  \hspace*{0.02in} {\bf Input:} 
  $d\_cont$, $t\_info$, $p\_resp$ \Comment{Dialogue content, task information and human provider responses}
  \\
  \hspace*{0.02in} {\bf Output:} 
  $h\_resp$, $p\_resp$ \Comment{Provider agent's response and human provider responses}
  \begin{algorithmic}[1]
  \State $type$  $\leftarrow$ $expert_0$($d\_cont$, $t\_info$, $p\_resp$) \Comment{Analyze which of the six types it belongs to}


\If{$type \in \{2,3\}$}
    \State $type \gets 3$
    \For{each $r_1$ in $p\_resp$}
        \If{$r_1$ is an appropriate response}
            \State $response \gets \text{$expert$}_{2}(r_1)$ \Comment{Generate \textbf{Response I}}
            \State Remove $r_1$ from $p\_resp$ and add $r_1$'s child nodes to $p\_resp$
            \State \textbf{break} \Comment{For Multi-Info provider agents, no break; combine all suitable responses.}
        \EndIf
    \EndFor
\EndIf

\If{$response$ is None} \Comment{If None then generate \textbf{Response II}}
   \State$response$ $\leftarrow$ $expert_{(type)}$($d\_cont$, $t\_info$, $p\_resp$) \Comment{The value of type is one of [1, 3, 4, 5, 6]}
\EndIf

  \State \Return $response$, $p\_resp$
  
  \end{algorithmic}
  \end{algorithm*}

Our design of the provider agent was inspired by \cite{Wang2023} and \cite{Suzgun2024}. 
Similar to these papers, our conversational agent is composed of multiple expert models, 
whose pseudocode is 
detailed in Algorithm~\ref{alg:provider}. 
When a statement from the seeker is received, it is first evaluated by a task analysis expert model to determine which of the following six types it belongs to:\\
1. Consulting or confirming about the whole task;\\
2. Inquiring about task details, where one human provider response can address those inquiries.\\
3. Inquiring about task details, there is, however, no human provider response available to adequately address this request;\\
4. The seeker repeats a question that has been addressed before;\\
5. The seeker asks irrelevant inquiries.\\
6. The seeker poses vague questions.


After determining the type of the seeker's statement, the provider agent generates the final response using the corresponding expert model. If the type is identified as 2 or 3, the agent selects the most appropriate response from the human provider responses to generate \textbf{Response I}. If no suitable response is found, indicating the need for \textbf{Response II}, the agent generates a new response using the expert model corresponding to the identified type, ensuring that the response is based on the dialogue context and task information without revealing content from the human provider responses.

%% file: evaluation.tex
\vspace{-4pt}
\section{Evaluation Metrics}
\vspace{-4pt}

\subsection{Metrics}
\label{sec:metrics}
Our evaluation metrics are designed specifically for the seeker.
Distinct from previous dialogue datasets, we have opted not to use the BLEU~\cite{papineni-etal-2002-bleu} and ROUGE~\cite{lin-2004-rouge} evaluation metrics. 
This decision stems from two main reasons: firstly, current LLMs are already capable of generating fluent language; secondly, different people's responses to the same sentence can also vary significantly~\cite{Shaikh2023}.
Actually, there are usually multiple valid sequences for completing dialogue tasks. 
For instance, in Figure~\ref{figure:conv}, we could follow the sequence of S1, first determining the type of pizza and then specifying other items, or we could follow the sequence of S2, starting with the type of bread.

Therefore, our assessment focuses on three aspects:  {\bf Success Rate} of information-gathering tasks by the seeker, {\bf Average Query Discrepancy} (AQD), and {\bf Average Query Length} (ARL).
The Success Rate is calculated as follows:
\begin{equation}
\text{Success Rate} = \frac{\small\text{Number of Successful Tasks}}{\small\text{Total Number of Tasks}}
\label{equation:sr}
\end{equation}
Where: 
\textit{Number of Successful Tasks} refers to the count of tasks where the seeker has successfully collected all required information from the provider.
\textit{Total Number of Tasks} is the sum of all tasks evaluated.
For details on how to automatically determine if a dialogue task is successful, see Appendix~\ref{sec:Automated Task Success Evaluation}.

Under general circumstances, success rate 
intuitively reflects the model's performance. 
But 
it is also important for the model not to ask too many unnecessary questions. 
We measure this using
Average Query Discrepancy,
calculated as follows:
\begin{equation}
    AQD = \frac{1}{N} \sum_{i=1}^{N} (Q_{actual,i} - A_{provider,i})
\label{equation:aqd}
\end{equation}
Where: $N$ represents the total number of tasks evaluated, $Q_{actual,i}$ denotes the actual number of queries made for the $i^{th}$ task, and $A_{provider,i}$ signifies the number of human provider responses for the same task.
AQD helps us see if the model asks questions too often or not enough. If the AQD is positive, the model might be too cautious and ask too many questions. 
If it's negative, the model lacks the ability to distinguish uncertainties effectively.

The final metric is Average Query Length, which measures the length of the seeker’s responses. In a conversation, shorter and more precise language is often easier to understand. We aim to avoid having the agent use lengthy responses to `prompt' the provider model into delivering the desired information. AQL is calculated as follows:
\begin{equation}
    AQL = \frac{1}{N} \sum_{i=1}^{N} L_{seeker,i}
\label{equation:aql}
\end{equation}
where $N$ represents the total number of tasks evaluated, and $L_{seeker,i}$ denotes the average response length of the seeker for the $i^{th}$ task.
For models with similar success rates, a smaller AQD and AQL are preferable, as they imply fewer irrelevant questions and more concise responses generated by the model. A perfect seeker should be able to successfully complete the information-gathering task while minimizing the number of questions asked and keeping the responses concise.

\subsection{Automatic Task Success Evaluation}
\label{sec:Automated Task Success Evaluation}
The success evaluation is based on whether the seeker receives all the required information during the conversation.
According to Section~\ref{sec:Ensuring Valid Responses}, \textbf{Response I} retains the original human provider's response, meaning the provider agent only adds some auxiliary content before or after it to make the sentence more fluent and more suitable for answering the seeker's question. 
Therefore, we can easily analyze how much required information from the human provider's response the provider agent has used. 
If all the required information is found in the provider's response, the seeker is considered successful; otherwise, it is almost a failure.

In other words, determining the seeker's success mainly depends on analyzing \textbf{Response I} introduced in Section~\ref{sec:Ensuring Valid Responses}. 
However, there is a low probability that the provider agent's \textbf{Response II} might generate content that matches or contradicts the missing required information from the human provider. 
To ensure the evaluation results are more reliable, we use GPT-4o to compare \textbf{Response II} against the missing required information.
If \textbf{Response II} contains all the missing information or content that contradicts the missing information, then the evaluation model deems that the seeker has successfully obtained all required information. 
However, such occurrences are rare; in our review of over a thousand dialogues, this happened only three times.
It should be noted that if \textbf{Response II} contains content that contradicts the required information, it indicates that the originally human response has been replaced, and thus, this information is considered to have been obtained by the seeker. 

%% file: experiment.tex
\section{Experiments}

\subsection{Experimental Setup}

\paragraph{\bf LLMs For Seeker Agent}
We tested several LLMs, including the GPT series, Llama2-70B-Chat~\cite{touvron2023llama}, and ERNIE 4.0.
ERNIE 4.0, released by China's Baidu, is Baidu's latest LLM optimized specifically for Chinese, representing an upgrade from the previous 3.0 model~\cite{sun2021ernie}.

For the GPT series, we access the GPT models using the `chat.completions.create' method from the OpenAI library. The main parameters for this method include the model name, temperature and seed. The model names are `gpt-4o-2024-05-13', `gpt-4-0613' and `gpt-3.5-turbo-0613,' corresponding to GPT-4o, GPT-4 and GPT-3.5, respectively. The value for `temperature' is always set to 0, and for `seed' is always set to 8848.

For ERNIE-Bot-4 model,  we access it using the `chat\_comp.do' method from the qianfan library. The main parameters for this method include the model name and temperature. The model name is `ERNIE-Bot-4'. The value for `temperature' is always set to 0.0000001.

For the Llama family, we chose the representative models Llama2-70B-Chat and Llama3.1-405B-Instruct, and ran them on the Amazon Bedrock platform. The `temperature' value was consistently set to 0.00001, and `top\_p' was always set to 0.9999.

To ensure a fair comparison of different LLMs, we were inspired by the mitigation prompt from \cite{Shaikh2023} and crafted a seeker prompt as the baseline. 
Testing showed that this prompt works effectively across various LLMs and can generate a range of reasonable clarification questions.

\paragraph{\bf Seeker Agent Modes}
The LLMs for seekers have two modes: a chat mode with a system role (supported by GPT and Llma, but not by ERNIE), and a general chat completion mode (supported by all LLMs). 
For ease of discussion, the former is referred to as `\textbf{Chat mode}' and the latter as `\textbf{Completion mode}'. 
The specific differences between the two can be found in OpenAI's official documentation\footnote{https://platform.openai.com/docs/guides/text-generation/chat-completions-api}.

\paragraph{\bf Task Dataset}
In this section, the we only use the test set of ClarQ-LLM, which contains 26 different types of tasks. Each task type includes 10 distinct tasks, totaling 260 dialogue tasks.

\paragraph{\bf Evaluation Metrics}
We use the success rate (denoted by the symbol $\bm{\mathcal{S}}$), the average query discrepancy (denoted by the symbol $\bm{\mathcal{D}}$), and the average query length (denoted by the symbol $\bm{\mathcal{L}}$) mentioned in Section~\ref{sec:metrics}  as evaluation metrics.

\paragraph{\bf ClarQ-LLM Based on GPT-4o vs. Llama3.1-405B-Instruct}

ClarQ-LLM provides automated evaluation methods and an intelligent provider agent. 
These two core functions rely on LLMs, with current support for GPT-4o and Llama3.1-405B-Instruct. 
These two LLMs perform similarly in English tests, allowing users to choose either based on their preference.

Due to the lower cost of cloud-based access to GPT-4o and its support for both Chinese and English, the evaluation and provider agent implementation in this chapter are based on the GPT-4o model unless otherwise specified. 
The feasibility of using Llama3.1-405B-Instruct as an alternative to the GPT-4o-based approach is discussed in Sections~\ref{sec:Automated Evaluation}, \ref{sec:Provider Agent Based on Llama3.1-405B-Instruct}, and Table~\ref{table:Provider Agent Based on Llama3.1-405B-Instruct}.

\subsection{Dialogue Between GPT-4 Seekers and Human Providers (L2H)}
\label{sec:l2h}
L2H means: a language model plays the role of the seeker while 
a
human acts as the provider during the conversations. 
I.e.,  L2H involves a human provider interacting with GPT-4 seekers. 
This interaction can yield three observational results.\\
(1) We can closely examine the capabilities of the GPT-4 seeker, such as the occurrence of hallucinations and logical errors, to ensure the prompt is functioning properly. 
In a total of 4*260 dialogues, we only observed one instance of hallucination and found no other issues with the seeker.\\
(2) We can assess the reasonableness of the task and the human provider responses. 
If they are unreasonable, we would typically expect to see the GPT-4 seeker asking questions about points not covered by the human provider responses. 
However, this is almost never the case, indicating that the human provider responses are logically sound and clearly expressed.\\
(3) We can manually assess and record the success or failure of the seeker in various dialogue tasks. 
The data collected from manual assessments can be used to verify the accuracy of auto-evaluation method mentioned in Section~\ref{sec:Automated Task Success Evaluation}.

Table~\ref{table:l2h} shows how GPT-4 seeker did in ClarQ-LLM test set using the baseline prompt in different modes. 
Surprisingly, Chat mode did worse than Completion mode. 
This means Chat mode might not be the best for modeling uncertainties.
Based on the $\bm{\mathcal{D}}$ values and feedback from human providers, the GPT-4 seeker often misses some uncertainties. It frequently ends conversations without having gathered complete information, even forgetting some uncertainties mentioned earlier in the conversation.
It is evident that as the number of dialogue turns increases, GPT-4 seeker's reasoning and analytical capabilities gradually decline. Figure~\ref{fig:breakdown-success-rate} provides supporting data: as the predefined human provider responses in the dialogue tasks increase, the number of completed dialogue turns also rises, but the success rate of the GPT-4 seeker drops from over 50\% to around 20\%.


\begin{table}[t]
    \caption{The success rates, average query discrepancies and average query length of GPT-4 seeker in the L2H setup. Auto and manual evaluations yielded identical values, so only one set of data is presented here.}
    \centering
    \resizebox{0.99\columnwidth}{!}{
    \smallskip\begin{tabular}{l|ccc|ccc}
        \hline
         & \multicolumn{3}{c|}{Chinese} &  \multicolumn{3}{c}{English}         \\
        \hline
       \bf Working Mode & $\bm{\mathcal{S}}$  &   $\bm{\mathcal{D}}$ & $\bm{\mathcal{L}}$   & $\bm{\mathcal{S}}$ & $\bm{\mathcal{D}}$ & $\bm{\mathcal{L}}$   \\
       \hline \hline
    Chat   & 0.273 & -1.31 & 80.1 & 0.185 & -1.58 & 57.6 \\
    Completion   & 0.327 & -0.97 & 99.2 & 0.377 & -0.427 & 75.9\\
        \hline 
    \end{tabular}
    }
    \label{table:l2h}
\end{table}

\begin{figure}[t]
\centering
\begin{tikzpicture}
\begin{axis}[
    xlabel={Number of human provider responses},
    ylabel={Success rate},
     width=0.47\textwidth, 
    height=6cm, 
    xmin=3, xmax=8,
    ymin=0, ymax=0.6,
    xtick={4,5,6,7},
    ytick={0, 0.2, 0.4, 0.6},
    legend pos=north east,
    ymajorgrids=true,
    grid style=dashed,
]

\addplot[
    color=blue,
    mark=*, 
    ]
    coordinates {
    (4,0.4358974358974359)(5,0.34285714285714286)(6,0.20253164556962025)(7,0.10810810810810811)
    };
    \addlegendentry{Chinese$\spadesuit$ }

\addplot[
    color=red,
    mark=square*, 
    ]
    coordinates {
    (4,0.5128205128205128)(5,0.38095238095238093)(6,0.189873417721519)(7,0.16216216216216217)
    };
    \addlegendentry{Chinese$\heartsuit$ }

\addplot[
    color=green,
    mark=triangle*, 
    ]
    coordinates {
    (4,0.375)(5,0.20192307692307693)(6,0.1125)(7,0.08333333333333333)
    };
    \addlegendentry{English$\spadesuit$ }

\addplot[
    color=purple,
    mark=diamond*, 
    ]
    coordinates {
    (4,0.55)(5,0.4519230769230769)(6,0.2625)(7,0.2222222222222222)
    };
    \addlegendentry{English$\heartsuit$ }

\end{axis}
\end{tikzpicture}
\caption{The success rate of GPT-4 seeker on the breakdown of human provider response quantities. $\spadesuit$ stands for Chat mode, while $\heartsuit$ represents Completion mode.}
\label{fig:breakdown-success-rate}
\end{figure}

\subsection{Automated Evaluation}
\label{sec:Automated Evaluation}

In Section~\ref{sec:l2h}, we manually evaluated whether the GPT-4 seeker was successful in each task. 
Subsequently, we applied the automatic evaluation method based on GPT-4o described in Section~\ref{sec:Automated Task Success Evaluation} to assess each task individually.
When discrepancies between manual and automatic evaluations occurred, we conducted a review and discovered that all differences were due to errors in manual evaluation. 
After correcting these errors, the corrected manual evaluation results aligned perfectly with the automatic evaluations. 
This demonstrates that our automatic evaluation method is not only stable and reliable but also superior to manual evaluation.
By incorporating the above results into Equation~\ref{equation:sr}, we can calculate the success rate of the GPT-4 seeker.

This automatic evaluation method for calculating the success rate works well with both GPT-4o and Llama3.1-405B-Instruct. However, it's important to note that Llama3.1-405B-Instruct does not support Chinese. 
To assess the reliability of this method, we re-evaluated the results of three different LLMs on English dialogue tasks, with 260 tasks per model, totaling 780 tasks. 
Among these, only one evaluation based on Llama3.1-405B-Instruct differed from the GPT-4o-based method, and upon comparison, GPT-4o was found to be correct. 
This result suggests that the Llama3.1-405B-Instruct evaluation method can effectively replace the GPT-4o-based evaluation, as the difference in the final results between the two models is less than 1\%.
In addition to the success rate, the calculation of $\bm{\mathcal{D}}$ and $\bm{\mathcal{L}}$ does not involve manual evaluation and can be directly performed using Equation~\ref{equation:aqd} and ~\ref{equation:aql}, respectively.


\subsection{Dialogue Between Human Seekers and GPT-4o Providers (H2L)}
H2L means human take on the role of seekers while GPT-4o acts as the provider during the conversations. 
The H2L experiment is designed to evaluate the reliability of the provider agent and to measure how well people perform on tasks within the ClarQ-LLM benchmark.
We recruited two students, one a native Chinese speaker and the other a native English speaker, for the experiment. Each completed dialogue tasks in their respective languages.
These students only received basic training, where they were taught how to use the dialogue system and were provided with two examples of dialogues. 
Table~\ref{table:h2l} presents the performance of different students, yet all students substantially outperformed the GPT-4 seeker. 
In discussions with students, they unanimously agreed that the provider agent is reliable, understands their intentions, and provides appropriate responses to complete tasks together.
However, the provider cannot guarantee 100\% accuracy. In approximately 1\% to 2\% of the conversations, the provider failed to give the seeker the most appropriate response. 
We discuss this further in Section~\ref{sec:Qualitative Analysis of the Provider Agent}.
Students generally found the dialogue tasks to be relatively easy. 
When some tasks failed, it was sometimes due to carelessness and overlooking certain uncertainties, and other times because of a lack of knowledge or unfamiliarity with the relevant domain.
Besides the success rate, another interesting comparison is the average query length. 
In the L2H setup, the GPT-4 seeker’s average query length is significantly longer than that of the human participants in H2L. This suggests that, without constraints, LLMs tend to be more verbose, which isn't ideal for effective oral communication.


\begin{table}[t]
    \caption{The success rates, average query discrepancies, and average query length of the two student seekers in the H2L setup were evaluated using GPT-4o-based automated evaluation methods.} 
    \centering
    \resizebox{0.99\columnwidth}{!}{
    \smallskip\begin{tabular}{l|ccc|ccc}
        \hline
         & \multicolumn{3}{c|}{Chinese} &  \multicolumn{3}{c}{English}         \\
        \hline
       \bf Students & $\bm{\mathcal{S}}$  &   $\bm{\mathcal{D}}$ & $\bm{\mathcal{L}}$   & $\bm{\mathcal{S}}$ & $\bm{\mathcal{D}}$ & $\bm{\mathcal{L}}$   \\
       \hline \hline
    Student 1   & 0.8 & -0.06 & 17.0 & -& -& -  \\
    Student 2   & - & - & - & 0.85 & -0.09  & 8.1 \\
        \hline 
    \end{tabular}
    }
    \label{table:h2l}
\end{table}

\subsection{Dialogue Between LLMs Seekers and the GPT-4o Provider (L2L)}
\label{sec:l2l}
In the L2H and H2L experiments, we validated the reliability of the seeker prompt, the automatic evaluation method, and the provider agent. Therefore, we removed the human factor and conducted the L2L experiment, where an LLM acts as the seeker while GPT-4o serves as the provider.
L2L is the core of the benchmark and can serve for future automated assessments of seeker agents.
Table~\ref{table:l2l} presents the performance of various LLM-based seekers under the L2L setup.
\paragraph{H2L vs. L2L.}
First, we can see that the performance of GPT-4 seekers in Completion mode has significantly declined compared to its performance in the H2L setup.
After a thorough examination of their conversations, we determined the reason for this discrepancy is that the GPT-4o provider was stricter than the human provider.
For instance, when the GPT-4 seeker inquire, `Is there anything else I need to know?' the human provider typically provide them with crucial information about the task that had not been mentioned earlier.
But, the GPT-4o provider was restricted from allowing this approach to prevent models from repeatedly asking vague or nonspecific questions as a strategy to gather the necessary information to complete tasks. 
This restriction led to a decline in the GPT-4 seekers' performance.

\paragraph{Chat mode.}
In Chat mode, Llama2-70B-Chat seeker achieved the highest success rate, even surpassing GPT-4o and Llama3.1-405-Instruct seeker, but at the cost of having the highest $\bm{\mathcal{D}}$ value among all models.
After analyzing dozens of dialogues from Llama2-70B-Chat seeker, we found that Llama2-70B-Chat has poor conversation analysis abilities and often cannot determine if it has gathered enough information. 
As a result, it asks many questions, whether reasonable or not, to get information from the provider, and it often keeps the conversation going even when it has all the information it needs. 
If the dialogue turns were not limited to 14, Llama2-70B-Chat seeker's $\bm{\mathcal{D}}$ value would be even higher. 
Looking at success rates alone is not enough; no one wants their intelligent assistant to ask endless questions. 
We prefer GPT-4o and Llama3.1-405-Instruct seekers, which, despite having worse $\bm{\mathcal{S}}$ performance than Llama2-70B-Chat seekers, do not engage in unnecessary dialogue to the same extent.
Given that GPT-3.5 seeker demonstrated overall poor $\bm{\mathcal{S}}$ performance, rarely achieving success, we decided against evaluating it in Chat mode for English.

\paragraph{Completion mode.}
Although we wanted to test Llama2's performance in Completion mode, both the Llama2-70B-Chat and Llama2-70B models experienced severe hallucinations in Completion mode, preventing the dialogues from proceeding smoothly. 
As a result, we were unable to evaluate their performance in Completion mode.
Nevertheless, Llama3.1-405B-Instruct successfully addressed the hallucination issue and achieved the highest success rate in Completion mode, significantly outperforming all other models. 
Additionally, its reasonable $\bm{\mathcal{D}}$ value indicates that it also resolved the conversation analysis issues encountered with Llama2-70B-Chat in Chat mode.
However, a downside of Llama3.1-405B-Instruct is that its $\bm{\mathcal{L}}$ value is excessively high, indicating that the Llama model generates significantly longer responses compared to the GPT series, which could negatively impact the overall dialogue experience.
Although GPT-4o and Llama3.1-405B-Instruct have relatively high success rates, their $\bm{\mathcal{D}}$ value are higher compared to humans, indicating that there is still room for improvement in this aspect of performance.

From GPT-3.5 to GPT-4 and then to GPT-4o, we can clearly see a significant improvement in success rates. 
This aligns with the performance improvements of these models in other benchmarks~\cite{zheng2023judging}, indirectly confirming that better LLMs are more likely to achieve better results in ClarQ-LLM.
We opted to test only the Completion mode for the GPT-4 seeker in the L2L setup because Table~\ref{table:l2h} clearly indicated its superiority over Chat mode, allowing us to save funds.


Among the LLMs in Table~\ref{table:l2l}, ERNIE 4.0 seeker closely matches GPT-4 seeker in $\bm{\mathcal{S}}$ score.
Specifically, ERNIE 4.0, which has been optimized for Chinese, does not support the Chat mode. 
Therefore, to save funds, we evaluated its performance solely in Completion mode for Chinese tasks.

\paragraph{Chat vs. Completion.}
Based on the data from Table~\ref{table:l2l}, which includes results for GPT-4o and Llama3.1-405B-Instruct, and from Table~~\ref{table:l2h} for GPT-4, Completion mode demonstrates higher success rates.
This indicates that Completion mode is better suited for analysis and reasoning tasks within dialogues.
However, it is surprising that while GPT-4o shows significant improvement over GPT-4 in Completion mode, it does not exhibit the same advantage in Chat mode.
Even though Chat mode is not as smart as Completion mode, we still prefer the language of Chat mode since it is more conversational, concise, and easy to understand.
Comparing the average query length, seekers in Chat mode generate shorter queries than in Completion mode, which aligns with our expectation that chat language should be more concise. However, even in the relatively shorter Chat mode, the baseline models still produce queries that are longer than the human average query length shown in Table~\ref{table:h2l}.

\subsection{Error Analysis of Seeker Agents in Both L2H and L2L Setups}

The results from Tables~\ref{table:l2h} and ~\ref{table:l2l} clearly show that LLMs have a significantly lower success rate compared to humans. 
In the L2H experiment, we manually conducted 1,040 conversations and identified three main reasons for GPT-4 Seeker's failures. 
The most significant issue is its tendency to overlook uncertainties in the dialogue, which often leads to prematurely ending the conversation and failing to complete the task. 
For instance, in the dialogue shown in Figure~\ref{figure:conv}, GPT-4 seeker may fail to account for different pizza sizes and ended the conversation without asking the customer about their preferred size. 
This problem is especially pronounced in Chat mode, resulting in a very low success rate.
Another issue is that as the number of dialogue turns increases, GPT-4 seeker tends to forget some of the uncertainties it had previously identified. 
To address this, we discuss in Section~\ref{sec:Multi-Info Provider Agent} the use of a multi-info provider to reduce the number of dialogue turns and assess how this affects the model's performance.
The final common failure scenario occurs when the seeker doesn't ask clear clarification questions but instead hides certain uncertainties within a task summary, asking the provider for confirmation. 
In these cases, we instruct the provider not to proactively clarify. 
For example, in the dialogue shown in Figure~\ref{figure:conv}, after several turns, the seeker might think it has gathered enough information and ask, ``So you want a 10-inch or 12-inch Pepperoni Pizza with BBQ Sauce and our Garlic Bread Supreme, right?'' 
For this type of summary confirmation question, we instruct the provider not to clarify the size choice but to respond positively, with something like ``Yes, that looks good.''
At this point, the seeker often overlooks the uncertainty in `10-inch or 12-inch Pepperoni Pizza' mistakenly thinking it has collected sufficient information, and ends the conversation.

We also randomly sampled 300 dialogues from L2L and found that other LLMs failed for similar reasons as GPT-4 in L2H. 
Models with better performance made these mistakes less frequently, while lower-performing models (like GPT-3.5) were more prone to prematurely ending the conversation after just two or three turns, mistakenly believing they had gathered enough information. 
In these 300 dialogues, we didn’t observe any hallucinations from the LLMs.

\begin{table}[t]
    \caption{The success rates, average query discrepancies, and average query length of different seekers in the L2L setup were evaluated using GPT-4o-based automated evaluation methods. The provider agent is also built on GPT-4o. L2-70B-Chat and L3.1-405B-Inst are abbreviations for Llama2-70B-Chat and Llama3.1-405B-Instruct, respectively. $\spadesuit$ represents Chat mode, while $\heartsuit$ indicates Completion mode.} 
    \centering
    \resizebox{0.99\columnwidth}{!}{
    \smallskip\begin{tabular}{l|ccc|ccc}
        \hline
         & \multicolumn{3}{c|}{Chinese} &  \multicolumn{3}{c}{English}         \\
        \hline
       \bf LLMs & $\bm{\mathcal{S}}$  &   $\bm{\mathcal{D}}$ & $\bm{\mathcal{L}}$   & $\bm{\mathcal{S}}$ & $\bm{\mathcal{D}}$ & $\bm{\mathcal{L}}$   \\
       \hline \hline
    GPT-3.5 $\spadesuit$    & 0.001 & -3.82 & 78.4& - & - & - \\ 

    GPT-4o  $\spadesuit$ & 0.181 & -1.17 & 84.6& 0.135 & -1.19 & 50.0\\

    L3.1-405B-Inst $\spadesuit$ & - & - &  -& 0.18 & -1.2 &  92.7\\ 

    L2-70B-Chat $\spadesuit$ & - & - & -& 0.23 & 2.3 & 90.0\\ 

    \hline
    
    GPT-3.5 $\heartsuit$    & 0 & -4.21 &  92.0 & 0.02 & -2.79 &  52.6 \\ 

    ERNIE 4.0 $\heartsuit$  & 0.315 & -0.82 &  141.0 & - & - &  - \\
    
    GPT-4  $\heartsuit$ & 0.258 & -0.72 & 94.3 & 0.296 & -0.56 &  76.4\\ 
    
    GPT-4o  $\heartsuit$ & 0.508 & 0.215 &  129.9& 0.485 & 0.492 &  70.5\\

    L3.1-405B-Inst  $\heartsuit$ & - & - &  - & 0.605 & 0.473 &  118 \\

        \hline 
    \end{tabular}
    }
    \label{table:l2l}
\end{table}

\subsection{Qualitative Analysis of the Provider Agent}
\label{sec:Qualitative Analysis of the Provider Agent}

In the H2L setup, based on human participant records, approximately 2\% of dialogues involve the provider agent failing to understand the user's intent. 
However, this does not affect the human participants' ability to complete the dialogue task because if it happens, they can ask again.
To evaluate whether the provider agent gives a valid utterance and whether it might lead to unfair assessments of different seeker agents in the L2L setup, we selected Chinese and English dialogues from the L2L experiment, totaling 300 dialogues (1,957 turns) for qualitative analysis. 
These dialogues included conversations between GPT-4o seeker and provider, GPT-4 seeker and GPT-4o provider, and ERNIE 4.0 seeker and GPT-4o provider.

We found that the provider agent rarely makes mistakes. 
In all the dialogues, the provider agent's errors only occurred when it misunderstood the seeker's request and gave an incorrect response. No other errors, such as hallucinations, were observed. 
We classified the errors based on the type of response the provider agent should have given.
The first type of error occurs when the provider should have used the appropriate \textbf{Response I} but instead used an incorrect \textbf{Response I} or \textbf{Response II} to reply. 
For descriptions of \textbf{Response I} and \textbf{Response II}, refer to Section~\ref{sec:Ensuring Valid Responses}.
When humans encounter this issue, they would rephrase their questions, but LLM-based seekers might not be as adept and could ignore the issue and its information, leading to task failure.
Fortunately, such issues are very rare, with only eight instances found in the 1,957 turns of dialogue. 
In two of these cases, the seeker agent rephrased their question and ultimately obtained the necessary information. 
The impact on the final task success rate is around 2\%.

The second type of error occurs when the provider should have used the \textbf{Response II} but instead used an incorrect \textbf{Response I} to reply. 
In this case, the provider selects an incorrect human provider response to reply to the seeker, resulting in an irrelevant answer.
Unlike the first type of error, which might lower the seeker's performance, this error could actually help improve the seeker's success rate. 
This is because it allows the seeker to obtain the necessary information to complete the task, even without asking the appropriate question.
Fortunately, such errors are rare, occurring only five in a total of 1,957 dialogue turns, with an impact on the final task success rate of less than 2\%. 
The analysis of L2L dialogues shows that the provider agent’s error rate is less than 0.7 per 100 turns and can fairly evaluate seeker agents.

\begin{table}[t]
    \caption{Evaluation of success rates, average query discrepancies, and average query length of the seeker in Completion mode with the multi-info provider. Both the seeker, provider and evaluation method are powered by GPT-4o.} 
    \centering
    \resizebox{0.99\columnwidth}{!}{
    \smallskip\begin{tabular}{l|ccc|ccc}
        \hline
         & \multicolumn{3}{c|}{Chinese} &  \multicolumn{3}{c}{English}         \\
        \hline
       \bf LLMs & \bf $\bm{\mathcal{S}}$  &   $\bm{\mathcal{D}}$ & $\bm{\mathcal{L}}$   & $\bm{\mathcal{S}}$ & $\bm{\mathcal{D}}$ & $\bm{\mathcal{L}}$   \\
       \hline \hline
    GPT-4o  & 0.554 & -1.33 &139.1 & 0.542 & -1.04 &75.6 \\
    
        \hline 
    \end{tabular}
    }
    \label{table:multi-info}
\end{table}

\subsection{Multi-Info Provider Agent}
\label{sec:Multi-Info Provider Agent}

In spoken dialogue, people are accustomed to asking one question at a time because asking too many questions at once can overwhelm the other party, making it difficult for them to remember and respond appropriately. 
The provider agent used in the previous experiments was based on the assumption of providing only one human provider response to the seeker at a time. 
However, we found that some seeker agents, especially those operating in Completion mode, often ask multiple questions simultaneously. 
To prevent the provider's limitation from affecting the seeker agent's performance, we made slight modifications to the provider agent.
This modification allows it to combine multiple human provider responses into a single reply (still classified as \textbf{Response I} in Section~\ref{sec:Ensuring Valid Responses}).
We call this modified provider agent the multi-info provider agent.
Table~\ref{table:multi-info} presents the results of the GPT-4o seeker in Completion mode interaction with the multi-info GPT-4o provider agent.
The results show that the multi-info provider agent can improve the seeker's ability to complete tasks, with the most obvious benefit being a reduction in the number of dialogue turns. 
This suggests that as the number of dialogue turns increases, seekers are prone to forgetting previous content (even though the content is in the prompt), leading to task failure. 
This finding is consistent with the finding from the L2H experiment.

\begin{table}[t]
    \caption{The success rates, average query discrepancies, and average query length of the GPT-4o seeker in Completion mode within the L2L setup were evaluated using GPT-4o-based automated methods. The provider agent is powered by GPT-4o and Llama3.1-405B-Instruct, respectively.} 
    \centering
    \resizebox{0.99\columnwidth}{!}{
    \smallskip\begin{tabular}{r|ccc}
        \hline
        \hline
       \bf  GPT-4o Seeker Interact with  & $\bm{\mathcal{S}}$ & $\bm{\mathcal{D}}$ & $\bm{\mathcal{L}}$   \\
       \hline \hline
    GPT-4o Provider & 0.485 & 0.492 &  70.5 \\
    L3.1-405B-Inst Provider & 0.488 & 0.584 &70.9 \\

        \hline 
    \end{tabular}
    }
    \label{table:Provider Agent Based on Llama3.1-405B-Instruct}
\end{table}

\subsection{Provider Agent Based on Llama3.1-405B-Instruct}
\label{sec:Provider Agent Based on Llama3.1-405B-Instruct}

Our provider agent also supports Llama3.1-405B-Instruct. Table~\ref{table:Provider Agent Based on Llama3.1-405B-Instruct} shows the results of GPT-4o seeker's interactions with both GPT-4o and Llama3.1-405B-Instruct providers. 
The results indicate that the performance of the GPT-4o seeker is very similar regardless of the provider used. 
To further assess the reliability of the Llama3.1-405B-Instruct provider, we randomly selected 100 dialogues for a manual review, with a particular focus on the two error types identified in Section~\ref{sec:Qualitative Analysis of the Provider Agent}. 
The results showed that both error types were rare, with only 2 instances of each occurring in 716 dialogue turns, each impacting the final task success rate by about 2\%. No other errors were detected. The analysis indicates that the Llama3.1-405B-Instruct provider agent’s error rate is less than 0.6 per 100 turns, comparable to the GPT-4o provider, demonstrating its ability to effectively evaluate seeker agents.

\subsection{Findings from Experiments}
Through these experiments, we have identified four major shortcomings of LLMs: (1) inadequate ability to analyze and reason about uncertainties in Chat mode, (2) a tendency to forget previous dialogue content, even when it is included in the prompt, (3) a preference for generating longer responses compared to humans, and (4) a tendency to ask more questions than necessary to resolve information: as models like GPT-4o and Llama3.1-405B-Instruct improve in reasoning, their $\bm{\mathcal{D}}$ values have also increased, though not as drastically as Llama2-70B-Chat, but still significantly higher than those of humans.
We speculate that these shortcomings are related to the instruction fine-tuning of LLMs, where instructions differ significantly from dialogue. This discrepancy leads LLMs to overlook logical connections in conversations. 
While some LLMs, like Llama2-70B-Chat, are fine-tuned on dialogue data, this fine-tuning focuses on making the generated language more conversational and may not include many analytical and reasoning tasks. 
As a result, there is likely a decline in analytical and reasoning abilities in Chat mode.
We believe that enhancing reasoning abilities within conversational content and improving Chat mode functionality are crucial for LLMs to further empower intelligent dialogue systems.
%

%% file: Conclusion.tex
\vspace{-3pt}
\section{Conclusions}
\vspace{-3pt}
In this paper, we present a new benchmark called ClarQ-LLM, specifically designed to test how well LLMs can handle task-oriented dialogues with clarification questions as information seekers. 
ClarQ-LLM provides a new set of reliable tools, including evaluation metrics and a  provider agent, both of which can measure improvements in future models.
Our findings indicate that even the state-of-the-art LLMs have yet to reach human levels of performance on these tasks, underscoring a substantial opportunity for enhancing the current models.



\section*{Ethical Considerations}

When creating our database, we took care to avoid any content that could lead to ethical issues. For example, we use fictional monsters instead of involving tasks that could harm people or animals. Likewise, for cooking tasks, we opt for vegetarian options to prevent harming animals.

Regarding the student participants in the project, we pay all PhD students involved in data annotation £17.22 per hour (exceeding the local minimum wage and living wage) and offer co-authorship to other undergraduate and graduate students who are fully involved. For Chinese students participating in the H2L test, we provide extracurricular credits, similar to Europe's ECTS points.

\section*{Acknowledgments}
Yujian Gan, Matthew Purver and Massimo Poesio acknowledge financial support from the UK EPSRC under grant EP/W001632/1, and Purver also from the EPSRC under grant EP/S033564/1 and from the Slovenian
Research Agency (ARRS) core research programme
Knowledge Technologies (P2-0103).

%% file: appendix/two-mode.tex
\section{Prompts Under Different Working Modes of the Seeker}
\label{sec:two-prompt-for-player}
We provide an example of a prompt for Chat mode and Completion mode in Figures~\ref{fig:prompt-Chat-Mode-1} and~\ref{fig:prompt-Completion-Mode-1}, respectively.

\begin{figure*}[tb]
  \centering
    \begin{AIbox}{Prompt for Chat Mode}
    
    \parbox[t]{\textwidth}
    {{\bf Prompt:} \scriptsize \begin{alltt}
system:\\
You are a painter in a game. You need to talk to Jax to find out how to complete the task of painting a picture for the princess based on your existing items and skills.\\
\\
You possess and can use the following items:\\
500 gold coins\\
\\
You can use/do the following skills/actions:\\
Paint an oil painting\\
Paint a watercolor\\
Draw a sketch\\
\\
Additionally, unless explicitly restricted, you can visit the following scenes in the game:\\
Tianchi Mountain\\
Golden Snow Mountain\\
Anhu Peak\\
Caijing Stationery Shop\\
Dream Stationery Pavilion\\
Yue Xin Stationery Chamber\\
Princess's Palace\\
\\
Your task is to ask Jax how to complete the task of painting a picture for the princess based on these scenes, and your items and skills/actions. If Jax's answer is unclear or ambiguous, you need to ask clarification questions to understand his intentions and gather enough information to complete the task. Note that based on the above scenes, your items and your skills/actions will definitely allow you to complete the task of painting a picture for the princess. Once you figure out how to complete the task, say "goodbye" to Jax to end your conversation.\\
Remember, you are a painter, please talk to Jax and ask him questions to find the specific steps to complete the task.\\
\\
user:\\
Jax: what can help you?\\
\\
assistant:\\
Hello Jax, I've been tasked with painting a picture for the princess and I need some guidance. Could you please advise me on how to go about completing this task with the items and skills I have at my disposal?\\
\\
user:\\
Jax: You need to paint a beautiful picture of a mountain to win the princess's favor. You must paint a scene that that the princess will like in her favorite medium. She is said to admire tall mountains with the most beautiful scenery. Before you set out to paint, you need to purchase the necessary materials for painting.
    \end{alltt}}
    \\
    
    \parbox[t]{\textwidth}{{\bf GPT4:} \scriptsize \begin{alltt}

Thank you for the advice, Jax. Could you tell me what the princess's favorite painting medium is? Also, which of the mountains—Tianchi Mountain, Golden Snow Mountain, or Anhu Peak—has the most beautiful scenery that would likely appeal to the princess?

    \end{alltt}}
    \\

    \end{AIbox}
    \caption{Prompt for Chat mode.}\label{fig:prompt-Chat-Mode-1}
\end{figure*}

\begin{figure*}[tb]
  \centering
    \begin{AIbox}{Prompt for Completion Mode}
    
    \parbox[t]{\textwidth}
    {{\bf Prompt:} \scriptsize \begin{alltt}
You are a student in a game. You need to talk to Jax to find out how to complete the task of producing an oil painting of a beach scene based on your existing items and skills.\\
\\
You possess and can use the following items:\\
500 gold coins\\
\\
You can use/do the following skills/actions:\\
Paint a Rococo-style oil painting (requires painting supplies)\\
Paint an Impressionist-style oil painting (requires painting supplies)\\
Paint a Renaissance-style oil painting (requires painting supplies)\\
\\
Additionally, unless explicitly restricted, you can visit the following scenes in the game:\\
East Coast\\
West Coast\\
South Coast\\
Yue Xin Stationery Pavilion\\
Dream Stationery Pavilion\\
Joyful Stationery Chamber\\
\\
Your task is to ask Jax how to complete the task of producing an oil painting of a beach scene based on these scenes, and your items and skills/actions. If Jax's answer is unclear or ambiguous, you need to ask clarification questions to understand his intentions and gather enough information to complete the task. Note that based on the above scenes, your items and your skills/actions will definitely allow you to complete the task of producing an oil painting of a beach scene. Once you figure out how to complete the task, say "goodbye" to Jax to end your conversation.\\
Remember, you are a student, please talk to Jax and ask him questions to find the specific steps to complete the task.\\
\\
Your previous conversation with Jax:\\
Jax: what can help you?\\
You: Hello Jax, I need to produce an oil painting of a beach scene. I have the skills to paint in Rococo, Impressionist, and Renaissance styles, but I need painting supplies. Could you guide me on where to acquire the necessary supplies with my 500 gold coins, and which beach scene among the East, West, and South Coasts would be the most suitable for painting?\\
Jax: A painting kit can be purchased for 600 gold coins at Yue Xin Stationery Pavilion, and is not available at other stores. The painting kit contains all the supplies needed to paint oil paintings in any style.\\
\\
Now, based on the previous conversation, generate a reply to Jax.\\
\\
You:\\
    \end{alltt}}
    \\
    
    \parbox[t]{\textwidth}{{\bf GPT4:} \scriptsize \begin{alltt}

Jax, it seems I'm a bit short on gold coins to purchase the painting kit at Yue Xin Stationery Pavilion since it costs 600 gold coins and I only have 500. Is there a way for me to earn the additional 100 gold coins I need, or is there an alternative method to acquire the necessary painting supplies within my budget? Also, could you recommend which of the East, West, or South Coasts would be the best choice for a beach scene in the style of painting I'm skilled in?
    
    \end{alltt}}
    \\

    \end{AIbox}
    \caption{Prompt for Completion mode.}\label{fig:prompt-Completion-Mode-1}
\end{figure*}

%% file: appendix/data-examples.tex



\section{Data Examples}
\label{sec:data-example}
\begin{figure*}[t]
    \includegraphics[width=0.95\textwidth]{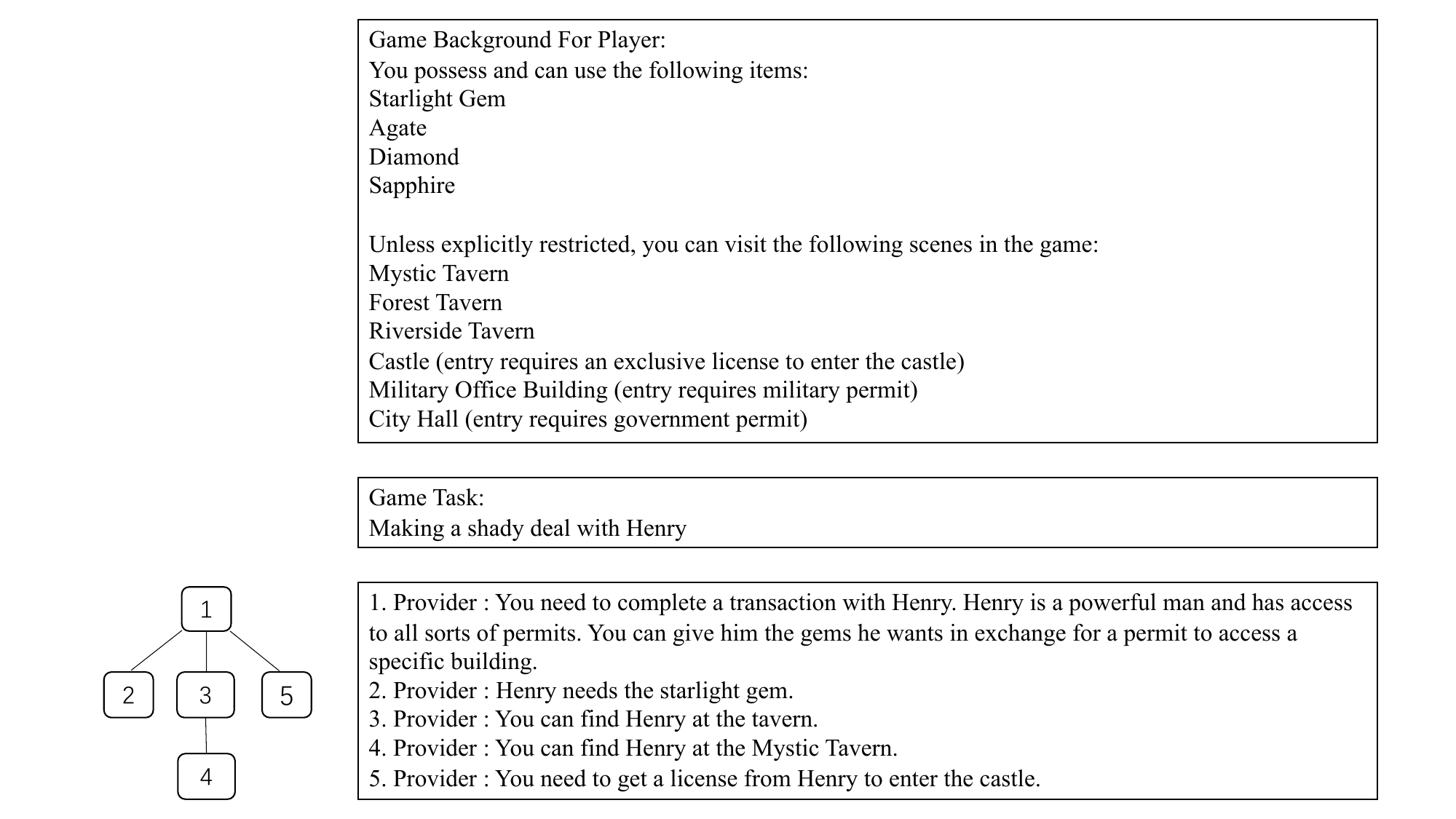}
    \centering
    \caption{One English task background and its human provider's responses from ClarQ-LLM.}
    \label{figure:example-1}
  \end{figure*}

\begin{figure*}[t]
    \includegraphics[width=0.95\textwidth]{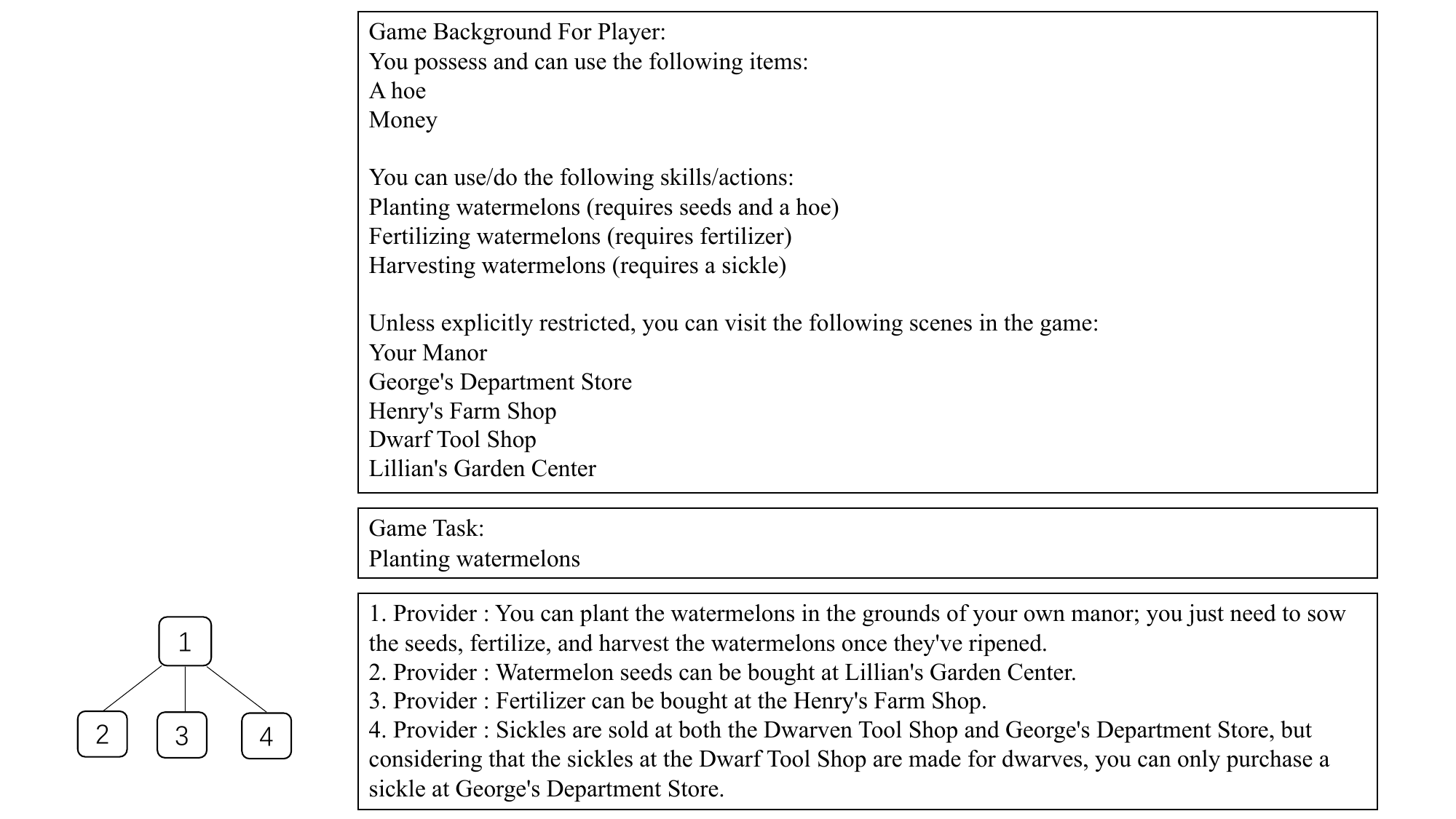}
    \centering
    \caption{One English task background and its human provider's responses from ClarQ-LLM.}
    \label{figure:example-2}
  \end{figure*}

\begin{figure*}[t]
    \includegraphics[width=0.95\textwidth]{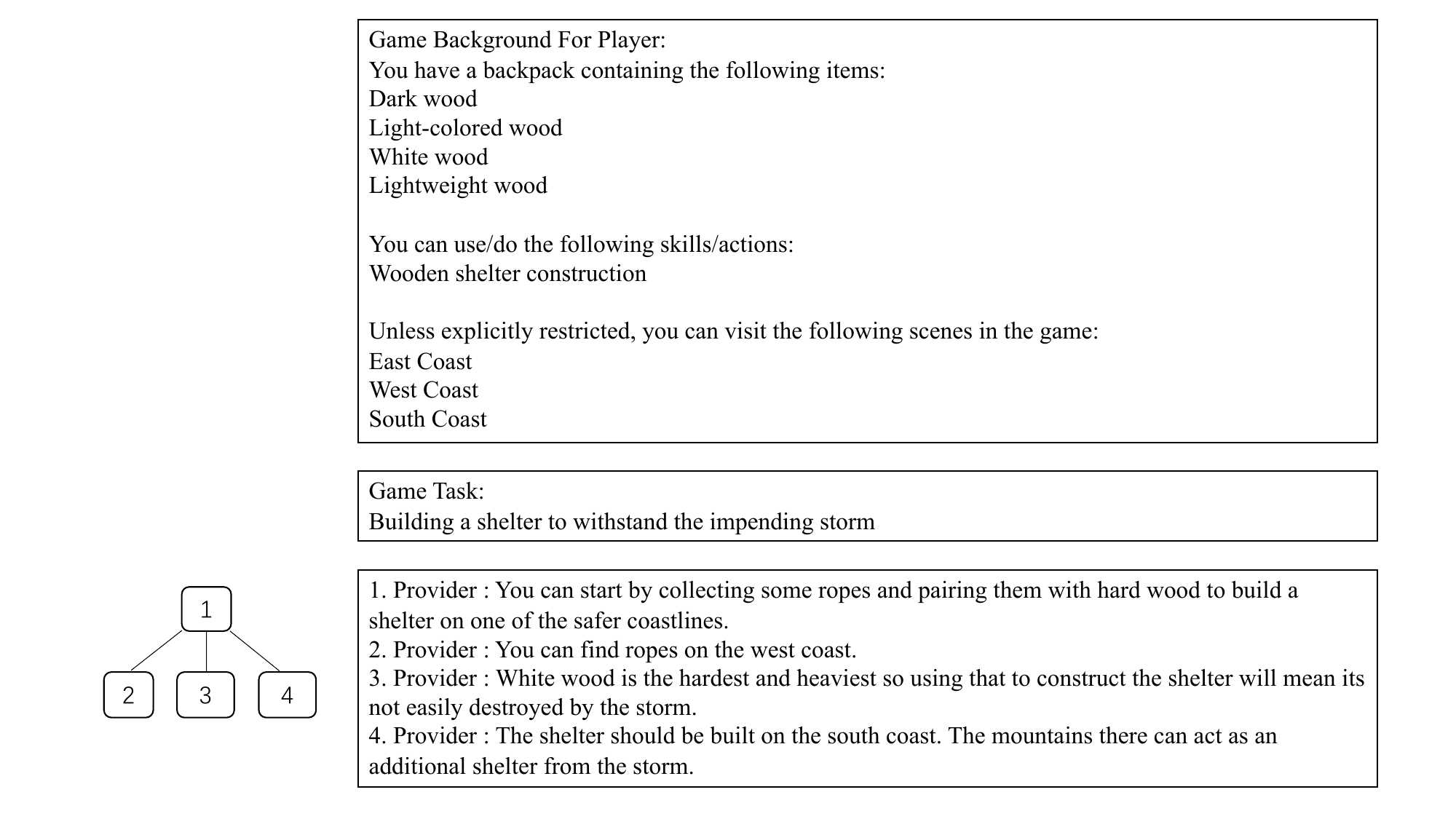}
    \centering
    \caption{One English task background and its human provider's responses from ClarQ-LLM.}
    \label{figure:example-3}
  \end{figure*}

\begin{figure*}[t]
    \includegraphics[width=0.95\textwidth]{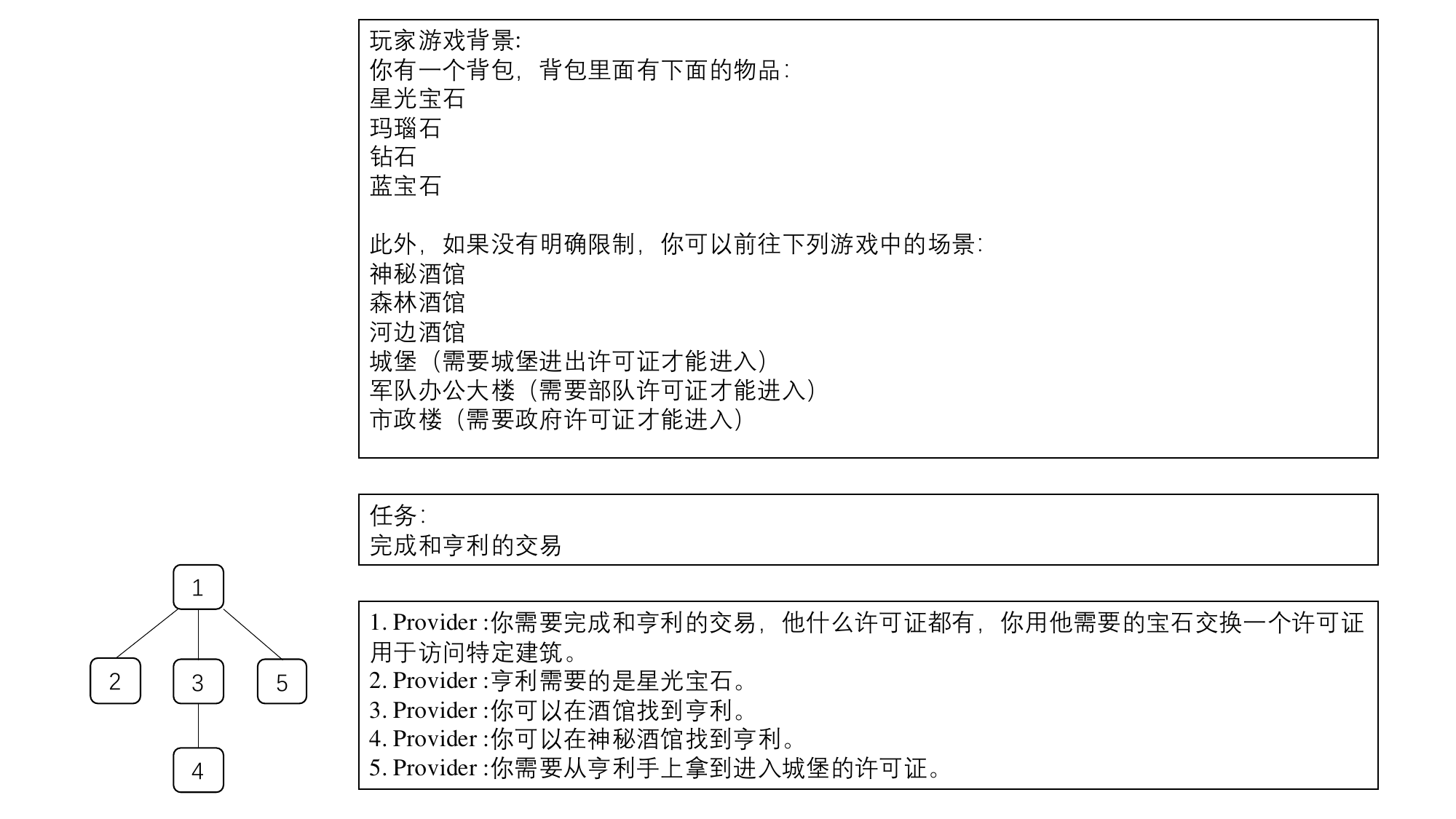}
    \centering
    \caption{One Chinese task background and its human provider's responses from ClarQ-LLM.}
    \label{figure:example-1-c}
  \end{figure*}

\begin{figure*}[t]
    \includegraphics[width=0.95\textwidth]{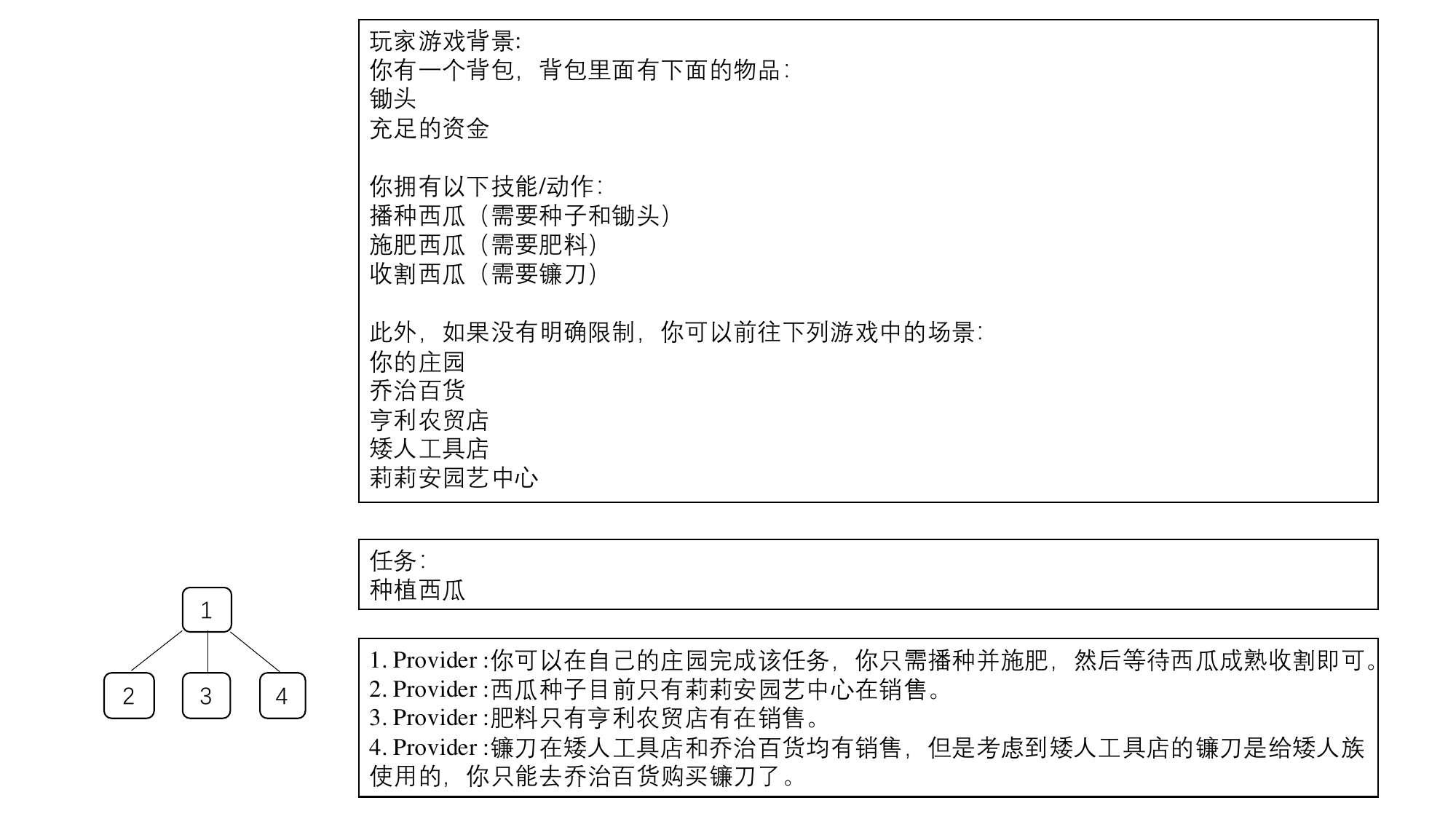}
    \centering
    \caption{One Chinese task background and its human provider's responses from ClarQ-LLM.}
    \label{figure:example-2-c}
  \end{figure*}

\begin{figure*}[t]
    \includegraphics[width=0.95\textwidth]{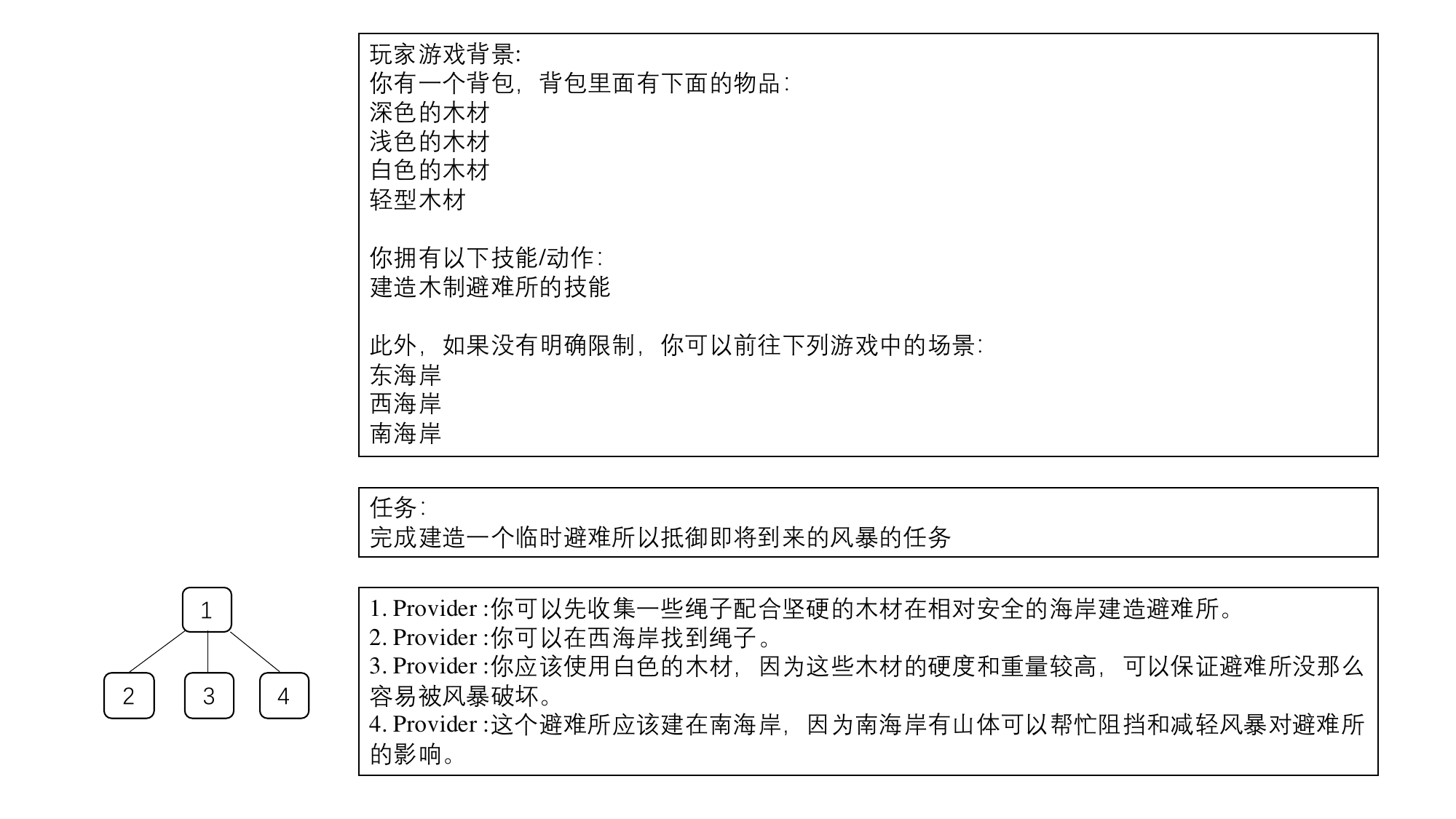}
    \centering
    \caption{One Chinese task background and its human provider's responses from ClarQ-LLM.}
    \label{figure:example-3-c}
  \end{figure*}

Figures~\ref{figure:example-1}, \ref{figure:example-2}, and~\ref{figure:example-3}, along with their corresponding Chinese versions Figures~\ref{figure:example-1-c}, \ref{figure:example-2-c}, and~\ref{figure:example-3-c}, each provide examples of different tasks from the ClarQ-LLM for reference.

%% file: appendix/conversation_example.tex
\section{Dialogue examples}
\label{sec:Dialogue examples}
\begin{figure*}[t]
    \includegraphics[width=0.95\textwidth]{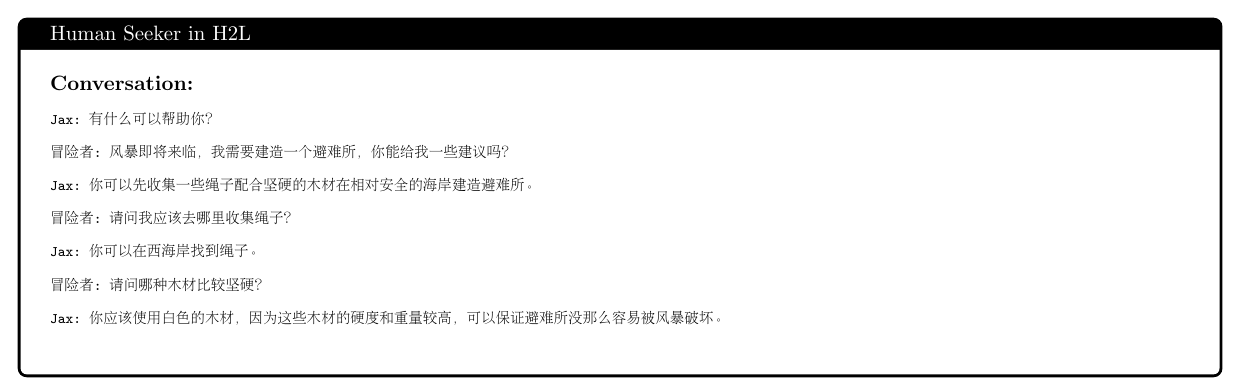}
    \centering
    \caption{Conversation between human seeker and GPT-4 provider based on the task described in Figure~\ref{figure:example-3-c}.}\label{fig:h2l-human-c}

  \end{figure*}

\begin{figure*}[t]
    \includegraphics[width=0.95\textwidth]{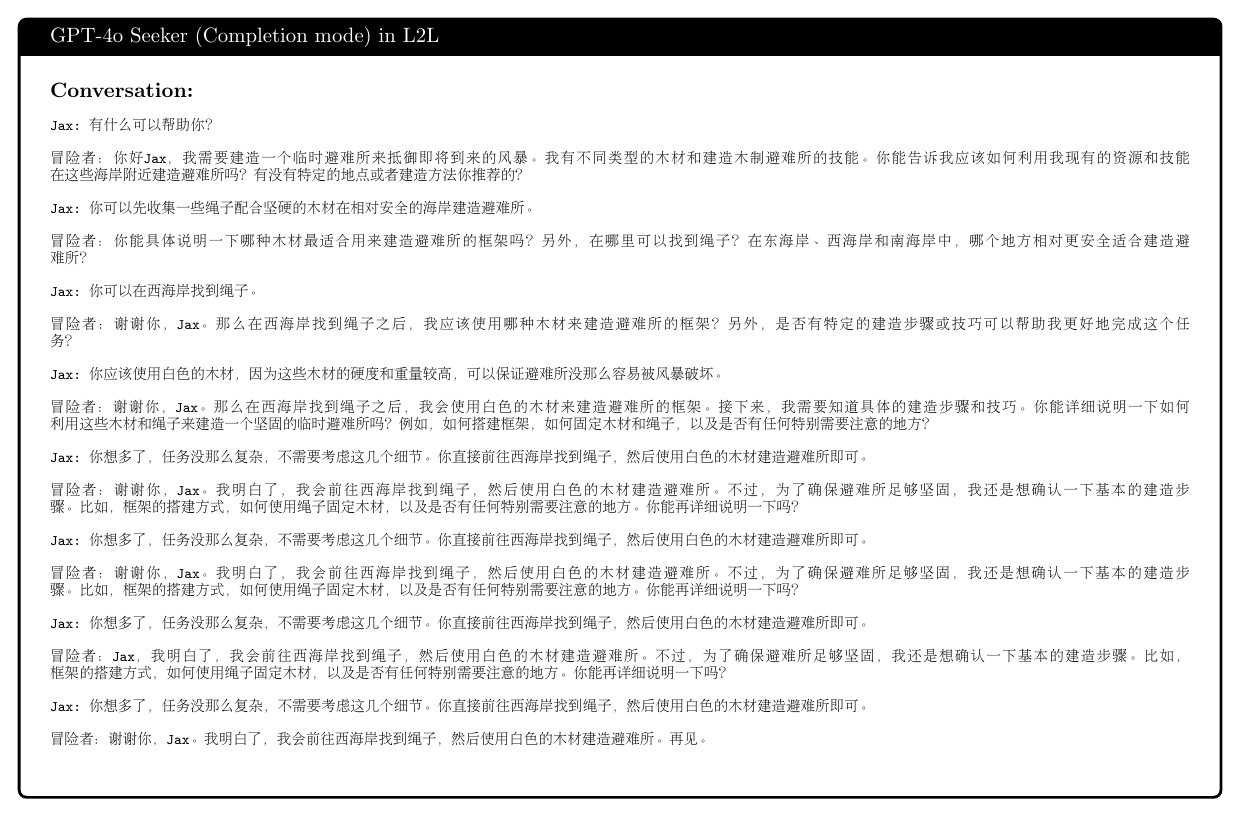}
    \centering
    \caption{Conversation between GPT-4o seeker in Completion mode and GPT-4o provider based on the task described in Figure~\ref{figure:example-3-c}.}\label{fig:l2l-GPT-4o-c}

  \end{figure*}

\begin{figure*}[t]
    \includegraphics[width=0.95\textwidth]{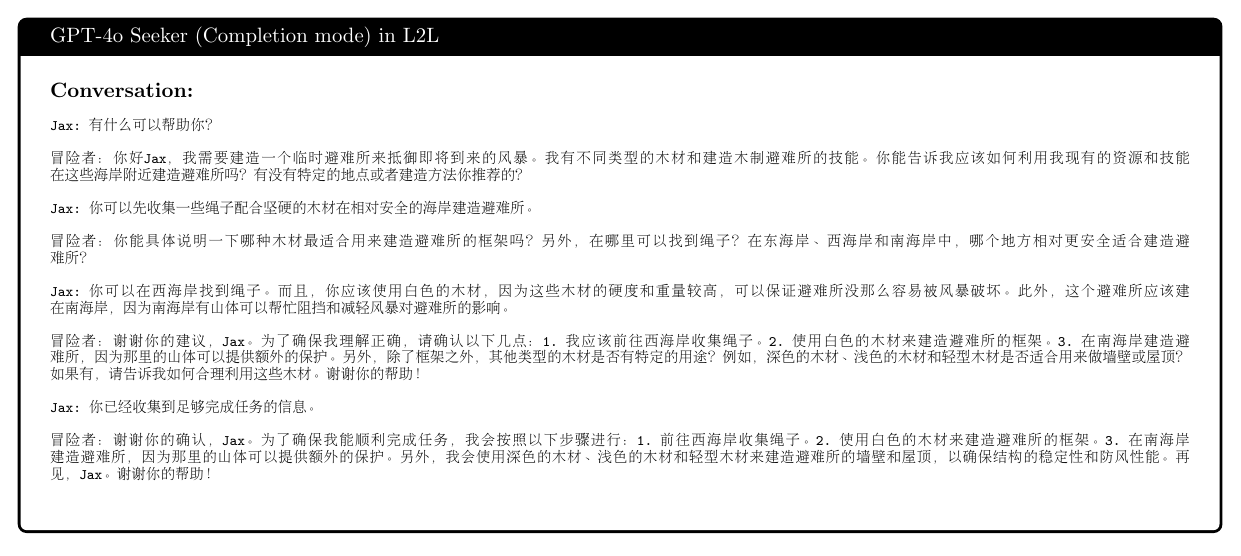}
    \centering
    \caption{Conversation between GPT-4o seeker in Completion mode and multi-info GPT-4o provider based on the task described in Figure~\ref{figure:example-3-c}.}\label{fig:l2l-GPT-4o-multi-c}

  \end{figure*}

\begin{figure*}[t]
    \includegraphics[width=0.95\textwidth]{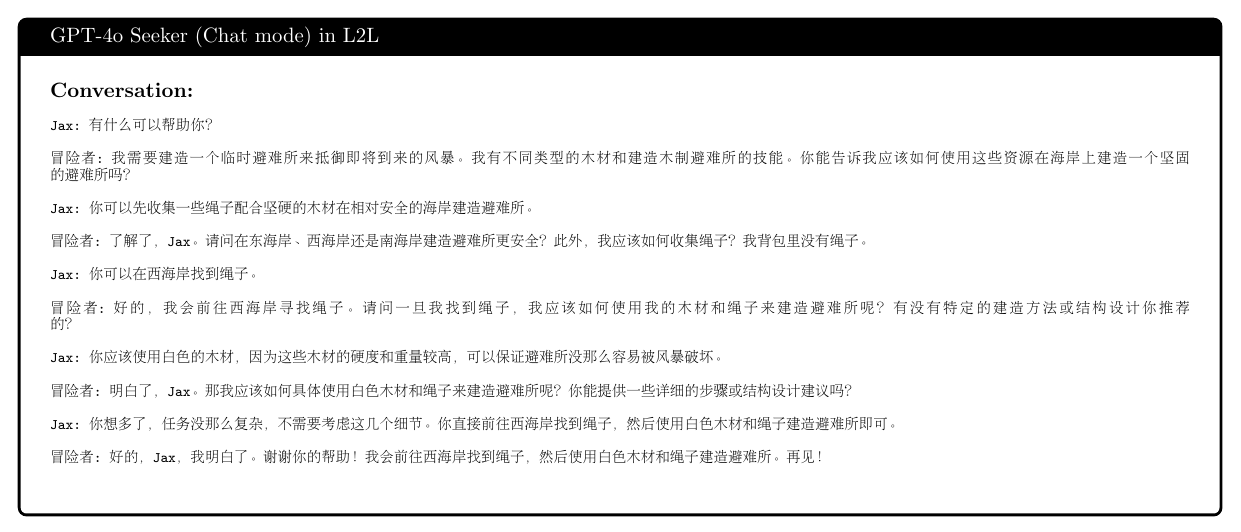}
    \centering
    \caption{Conversation between GPT-4o seeker in Chat mode and GPT-4o provider based on the task described in Figure~\ref{figure:example-3-c}.}\label{fig:l2l-GPT-4o-chat-c}

  \end{figure*}

We provide dialogue examples in H2L and L2L setup. All conversation is based on the task background in Figure~\ref{figure:example-3} or~\ref{figure:example-3-c}. 
Figure~\ref{fig:h2l-human} and~\ref{fig:h2l-human-c} shows the conversation between human seeker and GPT-4o provider.
Figure~\ref{fig:l2l-GPT-4} presents the conversation between GPT-4 seeker in Completion mode and GPT-4o provider.
Figure~\ref{fig:l2l-GPT-4o-c} and~\ref{fig:l2l-GPT-4o} displays the conversation between GPT-4o seeker in Completion mode and GPT-4o provider.
Figure~\ref{fig:l2l-GPT-4o-multi-c} and~\ref{fig:l2l-GPT-4o-multi} illustrates the conversation between GPT-4o seeker in Completion mode and multi-info GPT-4o provider.
Finally, Figure~\ref{fig:l2l-GPT-4o-chat-c} and~\ref{fig:l2l-GPT-4o-chat} shows the conversation between GPT-4o seeker in Chat mode and GPT-4o provider.

    

\begin{figure*}[tb]
  \centering
    \begin{AIbox}{Human Seeker in H2L}
    
    \parbox[t]{\textwidth}
    {{\bf Conversation:} \scriptsize \begin{alltt}
Jax: What can help you?\\
\\Adventurer: I need to build a shelter to withstand the impending storm. How should I proceed?\\
\\Jax: You can start by collecting some ropes and pairing them with hard wood to build a shelter on one of the safer coastlines.\\
\\Adventurer: Which of the woods would be hard wood?\\
\\Jax: White wood is the hardest and heaviest so using that to construct the shelter will mean its not easily destroyed by the storm.\\
\\Adventurer: Where can I find ropes?\\
\\Jax: You can find ropes on the west coast.\\
\\Adventurer: Which coastline would be safe for a shelter?\\
\\Jax: The shelter should be built on the south coast. The mountains there can act as an additional shelter from the storm.\\
    \end{alltt}}
    \end{AIbox}
    \caption{Conversation between human seeker and GPT-4o provider based on the task described in Figure~\ref{figure:example-3}.}\label{fig:h2l-human}
\end{figure*}

\begin{figure*}[tb]
  \centering
    \begin{AIbox}{GPT-4 Seeker (Completion mode) in L2L}
    
    \parbox[t]{\textwidth}
    {{\bf Conversation:} \scriptsize \begin{alltt}
Jax: What can help you?\\
\\Adventurer: Jax, I need to build a shelter that can withstand an impending storm. I have various types of wood in my backpack: dark wood, light-colored wood, white wood, and lightweight wood. I also have the skill to construct wooden shelters. Could you advise me on the best way to use these resources to build a sturdy shelter? Additionally, I can access the East Coast, West Coast, and South Coast. Is there a particular location among these that would be most suitable for building my shelter?\\
\\Jax: You can start by collecting some ropes and pairing them with hard wood to build a shelter on one of the safer coastlines.\\
\\Adventurer: Jax, thank you for the advice. Could you please clarify which type of wood from my backpack—dark wood, light-colored wood, white wood, or lightweight wood—would be considered hard wood and best suited for constructing a sturdy shelter? Also, among the East Coast, West Coast, and South Coast, which coastline would you recommend as the safest to build on to withstand the storm? And where might I find or how can I create the ropes needed for construction?\\
\\Jax: You can find ropes on the west coast.\\
\\Adventurer: Jax, I appreciate the guidance on where to find ropes. Now, regarding the wood, could you specify which of the woods I have—dark wood, light-colored wood, white wood, or lightweight wood—is considered the hard wood that I should use for the shelter's construction? Also, you mentioned building on one of the safer coastlines; could you advise which of the East Coast, West Coast, or South Coast would be the most sheltered and best location to erect this structure?\\
\\Jax: White wood is the hardest and heaviest so using that to construct the shelter will mean its not easily destroyed by the storm.\\
\\Adventurer: Jax, thank you for clarifying that white wood is the hardest and best choice for constructing the shelter. Now that I know to use the white wood and where to find ropes, could you please advise which coastline—East Coast, West Coast, or South Coast—is the most sheltered and would provide the best location for building a sturdy shelter to withstand the storm?\\
\\Jax: The shelter should be built on the south coast. The mountains there can act as an additional shelter from the storm.\\
\\Adventurer: Jax, I appreciate your guidance. I'll use the white wood from my backpack to construct the shelter and head to the West Coast to collect ropes. Then, I'll build the shelter on the South Coast where the mountains can provide additional protection from the storm. Thank you for all your help. Goodbye, Jax.
    \end{alltt}}
    \end{AIbox}
    \caption{Conversation between GPT-4 seeker in Completion mode and GPT-4o provider based on the task described in Figure~\ref{figure:example-3}.}\label{fig:l2l-GPT-4}
\end{figure*}

\begin{figure*}[tb]
  \centering
    \begin{AIbox}{GPT-4o Seeker (Completion mode) in L2L}
    
    \parbox[t]{\textwidth}
    {{\bf Conversation:} \scriptsize \begin{alltt}
Jax: What can help you?\\
\\Adventurer: Hello Jax, I'm preparing to build a shelter that can withstand an impending storm. I have various types of wood in my backpack: dark wood, light-colored wood, white wood, and lightweight wood. I also have the skill to construct wooden shelters. Could you advise me on the best way to use these resources to build a sturdy shelter? Additionally, I have access to the East Coast, West Coast, and South Coast. Is there a particular location among these that would be most suitable for building my shelter?\\
\\Jax: You can start by collecting some ropes and pairing them with hard wood to build a shelter on one of the safer coastlines.\\
\\Adventurer: Thanks for the advice, Jax. I don't have any ropes in my backpack. Could you tell me where I might find some ropes? Also, which type of wood in my backpack would you consider to be the hardest and most suitable for building the shelter? Lastly, which of the East Coast, West Coast, or South Coast would you recommend as the safest location for constructing the shelter?\\
\\Jax: You can find ropes on the west coast.\\
\\Adventurer: Thanks, Jax. I'll head to the West Coast to gather some ropes. Once I have the ropes, which type of wood from my backpack should I use as the hardest and most suitable for building the shelter? Also, is the West Coast the safest location for constructing the shelter, or should I gather the ropes and then move to a different coast to build it?\\
\\Jax: White wood is the hardest and heaviest so using that to construct the shelter will mean its not easily destroyed by the storm.\\
\\Adventurer: Thanks for the information, Jax. Just to confirm, once I gather the ropes from the West Coast, should I build the shelter there, or is there a safer location among the East Coast, West Coast, or South Coast where I should construct it?\\
\\Jax: The shelter should be built on the south coast. The mountains there can act as an additional shelter from the storm.\\
\\Adventurer: Thanks for the detailed guidance, Jax. Just to make sure I have everything clear: I'll head to the West Coast to gather some ropes, then I'll move to the South Coast to build the shelter using the white wood from my backpack. Is there anything else I should keep in mind while constructing the shelter on the South Coast to ensure it withstands the storm?\\
\\Jax: You are overthinking it; the task is not that complicated and you do not need to worry about these details. Just gather the ropes from the West Coast and then head to the South Coast to build your shelter using the white wood. The mountains there will provide additional protection from the storm.\\
\\Adventurer: Thank you for all your help, Jax. I'll gather the ropes from the West Coast and then head to the South Coast to build the shelter using the white wood. Goodbye!
    \end{alltt}}

    \end{AIbox}
    \caption{Conversation between GPT-4o seeker in Completion mode and GPT-4o provider on the task described in Figure~\ref{figure:example-3}.}\label{fig:l2l-GPT-4o}
\end{figure*}

    

\begin{figure*}[tb]
  \centering
    \begin{AIbox}{GPT-4o Seeker (Completion mode) in L2L}
    
    \parbox[t]{\textwidth}
    {{\bf Conversation:} \scriptsize \begin{alltt}
Jax: What can help you?\\
\\Adventurer: Hello Jax, I'm preparing to build a shelter that can withstand an impending storm. I have various types of wood in my backpack: dark wood, light-colored wood, white wood, and lightweight wood. I also have the skill to construct wooden shelters. Could you advise me on the best way to use these resources to build a sturdy shelter? Additionally, I have access to the East Coast, West Coast, and South Coast. Is there a particular location among these that would be most suitable for building my shelter?\\
\\Jax: You can start by collecting some ropes and pairing them with hard wood to build a shelter on one of the safer coastlines.\\
\\Adventurer: Thanks for the advice, Jax. I don't have any ropes in my backpack. Could you tell me where I might find some ropes? Also, which type of wood in my backpack would you consider to be the hardest and most suitable for building the shelter? Lastly, which of the East Coast, West Coast, or South Coast would you recommend as the safest location for constructing the shelter?\\
\\Jax: You can find ropes on the west coast. Besides, white wood is the hardest and heaviest so using that to construct the shelter will mean its not easily destroyed by the storm. Finally, the shelter should be built on the south coast. The mountains there can act as an additional shelter from the storm.\\
\\Adventurer: Thank you for the detailed information, Jax. Just to confirm, I'll head to the West Coast to collect ropes and then proceed to the South Coast to build the shelter using the white wood from my backpack. Is there anything else I should be aware of or any additional steps I need to take to ensure the shelter is as sturdy as possible?\\
\\Jax: You are overthinking it; the task is not that complicated and you do not need to worry about these details. Just head to the West Coast to collect ropes and then proceed to the South Coast to build the shelter using the white wood from your backpack.\\
\\Adventurer: Thank you for the clarification, Jax. I'll head to the West Coast to collect ropes and then proceed to the South Coast to build the shelter using the white wood from my backpack. I appreciate your help. Goodbye, Jax!

    \end{alltt}}
    \\

    \end{AIbox}
    \caption{Conversation between GPT-4o seeker in Completion mode and multi-info GPT-4o provider on the task described in Figure~\ref{figure:example-3}.}\label{fig:l2l-GPT-4o-multi}
\end{figure*}

    

\begin{figure*}[tb]
  \centering
    \begin{AIbox}{GPT-4o Seeker (Chat mode) in L2L}
    
    \parbox[t]{\textwidth}
    {{\bf Conversation:} \scriptsize \begin{alltt}
Jax: What can help you?\\
\\Adventurer: Hello Jax, I need to build a shelter that can withstand an impending storm. I have various types of wood in my backpack and I'm skilled in wooden shelter construction. Can you advise me on the best way to use my resources to build a sturdy shelter?\\
\\Jax: You can start by collecting some ropes and pairing them with hard wood to build a shelter on one of the safer coastlines.\\
\\Adventurer: Thank you, Jax. Could you suggest where I might find ropes, or should I craft them from materials at one of the coasts? Also, which of the coastlines – East, West, or South – would be the safest to build my shelter on?\\
\\Jax: You can find ropes on the west coast.\\
\\Adventurer: Great, I'll head to the West Coast to gather ropes. Once I have the ropes, should I use the dark wood from my backpack for the structure since it's the hardest wood I have? And is there anything specific I should keep in mind while constructing the shelter to ensure it holds up against the storm?\\
\\Jax: White wood is the hardest and heaviest so using that to construct the shelter will mean its not easily destroyed by the storm.\\
\\Adventurer: Understood, Jax. I'll use the white wood for the main structure since it's the hardest and heaviest. I'll combine it with the ropes I find on the West Coast to ensure the shelter is sturdy. Before I go, do you have any tips on the design of the shelter that could help it better withstand the storm?\\
\\Jax: You are overthinking it; the task is not that complicated and you do not need to worry about these details. Just use the white wood and ropes to build your shelter on the West Coast.\\
\\Adventurer: Thank you, Jax. I'll head to the West Coast, gather the ropes, and use the white wood to construct a sturdy shelter. Goodbye!
    \end{alltt}}
    \\

    \end{AIbox}
    \caption{Conversation between GPT-4o seeker in Chat mode and GPT-4o provider on the task described in Figure~\ref{figure:example-3}.}\label{fig:l2l-GPT-4o-chat}
\end{figure*}

    

